\documentclass{article}
\usepackage[T1]{fontenc}
\PassOptionsToPackage{table}{xcolor}
\usepackage{preprint,times}

\usepackage{graphicx}
\usepackage{xcolor}
\usepackage{booktabs}
\usepackage{algorithm}
\usepackage{algpseudocode}
\usepackage{wrapfig}
\usepackage{amssymb}
\usepackage{hyperref}
\usepackage{url}
\definecolor{lightblue}{RGB}{204,229,255}
\definecolor{lightgreen}{RGB}{204,255,204}

\usepackage{amsmath,amsfonts,bm}

\def\eqref#1{equation~\ref{#1}}

\def\1{\bm{1}}

\DeclareMathAlphabet{\mathsfit}{\encodingdefault}{\sfdefault}{m}{sl}
\SetMathAlphabet{\mathsfit}{bold}{\encodingdefault}{\sfdefault}{bx}{n}

\newcommand{\parag}[1]{\vspace{-3mm}\paragraph{#1}}
\newcommand{\ours}[0]{{\it AnchorGen}}

\title{AnchorGen: Anchored Optimization for
\\
Customizable Generative 3D Design}

\author{%
Hantao Zhang,~~Oliver Heinimann,~~Jieke Wu,~~Yingxuan You,\\
\textbf{Emilien Seiler,}~~\textbf{Pascal Fua}\\
CVLab, EPFL, Switzerland
}

\newcommand{\encd}{{\mathrm{Enc}}}
\newcommand{\decd}{{\mathrm{Dec}}}

\begin{document}
\maketitle

\begin{abstract}

Engineering design often starts from a 2D sketch that fixes style and proportions, yet the subsequent 3D shape optimization relies on learned generative priors to keep the geometry valid. However, these priors are agnostic to the sketch: while they admit a valid design by correcting a drifted proposal back to its training distribution, they often correct it towards the high-density region, ignoring the specified design. We introduce \emph{AnchorGen}, a rectified-flow framework trained unconditionally on the concatenated shape and sketch latents of paired data. The learned manifold represents the joint distribution of shape-sketch pairs, so constraining the sketch component restricts the iterate to the sub-manifold of shapes consistent with a target style. Since training employs no conditioning signal, the constraint is imposed at inference: gradient descent optimizes the shape latent to minimize a differentiable drag surrogate, while constraining the sketch latent to remain close to the target sketch via a token-wise cosine penalty. A single model thereby supports design-preserving optimization, dimensionally explicit design edits, and sketch-only synthesis.

\end{abstract}

\vspace{-5mm}
\begin{figure}[h!]
    \centering
    \includegraphics[width=1\linewidth]{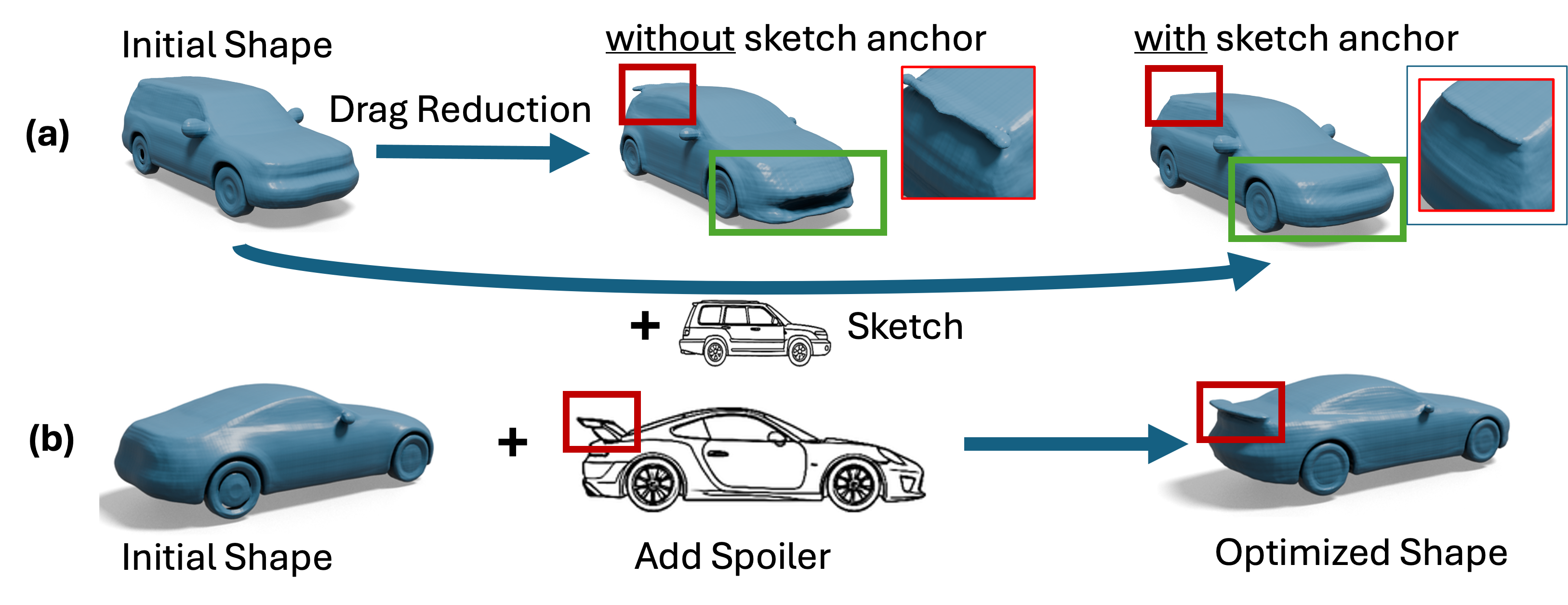}
   \vspace{-8mm}
    \caption{
    \small {\bf Sketch-anchored shape optimization.} \ours{} reduces the aerodynamic drag of a 3D vehicle while anchoring the optimization to a 2D sketch that encodes the design intent.
    (a)~\textbf{Preserving a design.} Reducing the drag by $10\%$ without a sketch anchor, as in FCSO~\citep{Seiler26a}, changes the vehicle's identity: the front end is restyled (green) and an unintended rear wing appears (red). With the sketch of the initial shape as an anchor, \ours{} reaches the same drag reduction and preserves both regions.
    (b)~\textbf{Modifying a design.} Adding a spoiler to the sketch (red) produces an optimized shape with a more aerodynamic spoiler, while leaving the rest unchanged.
    }
    \label{fig:anchorgen-teaser}
    \vspace{-1mm}
\end{figure}

\section{Introduction}
\label{sec:introduction}

Many 3D engineering design workflows start with a simple 2D sketch: A few contour lines often suffice to fix the proportions of a 3D shape together with the styling cues that carry its brand identity. This enables a designer to explore several potential directions in the time elsewise required to model a single one directly in 3D~\citep{Tovey23,Lee25c}. This makes sketching a natural medium for early-stage brainstorming and exploration before committing time to full geometry. Industrial pipelines for CAD~\citep{yang2026sketch2arti}, product design~\citep{cordier2013inferring}, and automotive styling~\citep{nozawa20223d} are organized accordingly, proceeding from sketching to {3D geometry and then to refinement.

This has spurred considerable interest in lifting 2D sketches to 3D geometry, and large sketch- and image-conditioned generators can now handle complex geometries~\citep{nozawa20223d,zheng2023locally,li2025meshpad}. However, in engineering design, the initial 3D shape is not the deliverable, but only the starting point of an optimization against a physical objective, such as minimizing aerodynamic drag or maximizing structural strength~\citep{You26a,Vatani25a}.
In most current pipelines~\citep{zhan2025physiopt,Vatani25a}, the sketch serves only to initialize the optimization. Its influence does not extend beyond the starting point, as the subsequent trajectory is determined by the physical objective and by the prior that constrains the design to remain valid.
In modern implementations~\citep{Seiler26a,You26a}, the optimization is often performed with respect to latent vectors~\citep{chen2025dora,xiang2025structured,Zhang23d} that parameterize the shapes, whereas the physical objective is evaluated in the explicit shape space, either by CFD~\citep{jasak2007openfoam} with an adjoint method~\citep{Othmer14} or by a surrogate model~\citep{Wu24a} trained on a set of exemplars. Once the optimization starts, the latent vector can easily move away from the manifold of vectors representing valid shapes, resulting in implausible ones and meaningless values of the objective function, either because CFD fails or because the surrogate models are not trained for such shapes. This is referred to as {\it drift} in~\citep{xiang2025structured,Seiler26a} and can be mitigated by projecting the candidates back into the region in which the models are valid, for example using guided generative algorithms~\citep{Lugmayr22,Meng22c, Seiler26a}. However, this amounts to using a prior trained on shapes alone to project onto the aggregate shape distribution. Even when the projection is guided by the physical objective, the correction itself is still applied by the aggregate shape prior, and thus favors generic shapes. In other words, the very mechanism that keeps the optimization valid is also what erases what the sketch specified: Such a prior is agnostic to design intent, it can certify that a design is valid, not determine which valid design it should be. An optimizer asked to reduce the drag of a pickup truck may therefore succeed by returning a sedan or, as shown in Fig.~\ref{fig:anchorgen-teaser}(a), add an unintended rear wing while restyling the front bumper into that of a sports-car.

What is missing is an interface for design intent. Parameterized deformation and learned parameter-controlled generators such as LAMP~\citep{Nehme25} provide one through explicit control variables~\citep{Zhang26b,Nehme25}, but they are bounded by a hand-designed parameterization~\citep{Othmer14,Masters17} and tied to a few scalars, which makes them ill-suited to intent expressed as a roofline, a silhouette, or a tail treatment. Diffusion-based 3D models~\citep{Seiler26a} are not subject to this restriction, since they learn the design space from data, but they come with no design interface in its place. The sketch that started the pipeline is a natural candidate for that interface~\citep{Tovey23}: The side view of a car captures features such as the roofline and the rear slant, which give the car its distinct look but can also severely affect drag.

In this paper, we turn the sketch into such an interface by making it part of the prior rather than an input to it. Our \ours{} framework, shown in Fig.~\ref{fig:framework}, first learns latent representations for shapes and sketches, and then trains a single flow matching model~\citep{Lipman22} over the \emph{concatenation} of the two, thereby learning the joint distribution of shapes and sketches. At inference time, we minimize the physical objective by steering the sampling trajectory of the flow:
We optimize the flow's intermediate states so that the decoded shape minimizes a surrogate estimate of the objective, while a penalty keeps the sketch component of the pair close to the latent vector of a target sketch in terms of cosine distance. Every candidate therefore remains a sample of the joint prior, and the penalty narrows that prior from the manifold of valid shapes to the subset consistent with the target. As a result, corrections that would otherwise pull a drifting design toward generic shapes of the data now pull it toward that subset. The sketch acts as an \emph{anchor} that persists throughout the optimization. Training is fully unconditional: Both modalities are noised and denoised together, so the model learns their correspondence instead of a conditional of one given the other. A conditional model must commit to its control interface before training, whereas ours makes it possible to choose what to constrain, how strongly, and alongside which objective without additional training.

Our contributions are therefore as follows:
\begin{itemize}

    \vspace{-2mm}
    \item \textbf{A prior over shape-sketch pairs.} We train a single unconditional flow over the joint latent of a shape and its sketch, which turns the sketch from an initialization into a constraint that persists throughout the optimization.

    \vspace{-2mm}
    \item \textbf{Anchoring at inference time.} We minimize a physical objective along this flow while constraining the sketch component to remain close to a target. The manifold correction is thereby directed to the sub-manifold of shapes consistent with the target sketch, instead of the high-density region of the shape distribution.
    \vspace{-1mm}
    \item \textbf{A single model for three design tasks.} Because the anchor is imposed only at inference, a single trained model supports three different usages; optimization anchored to an \emph{existing} sketch, optimization anchored to an \emph{edited} sketch that prescribes a targeted design change, and \emph{synthesis} from noise anchored to a sketch alone.

\end{itemize}
We validate \ours{} on two complementary benchmarks. ShapeNet~\citep{Chang15} spans diverse vehicle types, enabling evaluation of whether optimized shapes remain within the original shape distribution against prior-based methods. DrivAerNet++~\citep{elrefaie2024drivaernet++} provides explicit design parameters, enabling quantitative evaluation of how precisely customized designs are realized, with comparisons to conditioning-based methods. \ours{} achieves the best results across all metrics on both benchmarks and further generalizes to unseen sparse freehand sketches.

\section{Related Work}
\label{sec:related}

\subsection{Controllable Generative Models for 3D Shapes}
\label{sec:genmodels}

Controllable 3D generative models, most of which are now diffusion-based, combine a learned shape prior with user-specified evidence~\citep{Wang25a,Wei26a}.
How that evidence enters divides existing work into two families. Input-conditioned methods learn the constraint
at training time: The prescribed signal is encoded and injected into the denoiser or velocity field at every sampling
step~\citep{Ho22a,cheng2023sdfusion,xiang2026native,Lee25b,cen2026text}. Such methods are efficient at inference, but the control signal is fixed at training time and cannot be optimized.
Test-time steering methods
instead leave the prior untouched and steer its inference trajectory by minimizing a
differentiable loss evaluated on the running clean
estimate~\citep{Dhariwal21a,Chung22a,song2023loss,BenHamu24}. Such methods accommodate any differentiable objective without retraining and allow the constraint itself to be optimized, but require iterative sampling.

In industrial 3D design, the generated shape must additionally satisfy physical objectives, such as minimizing aerodynamic drag, rather than only match a desired appearance. Performance-driven workflows therefore combine both mechanisms. They first use
input conditioning to decode a latent vector into a 3D shape that conforms to
the design intent, and then optimize a physical loss with respect to that
vector~\citep{Morita24,Vatani25a}. Preserving the shape prior during this second
stage can be done in two main ways. The first~\citep{zhan2025physiopt} relies on a compact latent space~\cite{Park19c} to keep optimization near the shape manifold, but this constraint weakens in high dimensions~\cite{chen2025dora}, where the curse of dimensionality makes optimization more prone to drifting toward meaningless latents. The second is to let the gradient
of the physical loss drive the generation, allowing the latent vector to drift
away from the manifold of vectors that correspond to meaningful
shapes~\citep{Liu26a}, and to periodically re-noise and de-noise the candidates
to project them back onto that manifold~\citep{Lugmayr22,Meng22c,Li26a,You26a,Seiler26a}.
The difficulty is that this projection is not guaranteed to return high-quality
candidates.

We overcome these limitations by jointly modeling the shapes and the corresponding sketches using a single latent vector for both. Such joint modeling has been explored before for shape generation purposes~\citep{cheng2022cross,bao2023one,li2025omniflow}. It has also been tried for image segmentation~\citep{Zhang25b} and other discriminative tasks~\citep{Zhang26a}.  However, in all these methods, a  variable is observed and the other is produced. Thus, the design focuses on  architectures that support this  exchange. In contrast, we optimize the shape and sketch jointly to minimize a physical loss, while constraining the sketch to remain close to a target. The sketch thereby acts as an \emph{anchor} that restricts the search to designs consistent with the target sketch.

\subsection{Sketch-Based 3D Design}
\label{subsec:sketch3d}

Sketches are a key medium for early-stage industrial design: A few contour and character lines simultaneously fix the overall proportion of a shape and the styling details that carry brand identity.
A designer can produce them much faster than a CAD parameterization, a photorealistic rendering, or a 3D scan~\citep{Tovey23,Lee25c,huang2026handmade}.
However, design is only one of the applications that have motivated work on sketches, and the algorithms that lift a 2D drawing to a 3D shape have largely been developed as solutions to a general-purpose inverse problem rather than as design tools.
Early work addressed the inherent ambiguity of that problem either by assuming
that drawn lines correspond to specific shape features~\citep{Malik89,Igarashi06}, or by restricting the class of representable 3D shapes~\citep{Leclerc92,Lipson96,cordier2013inferring,jung2015sketching,Dvorovzvnak20}.
Deep learning brought more expressive models and better generalization~\citep{Lun17a,Delanoy18,Jin20b,nozawa20223d}.More recently, diffusion models have been conditioned on sketches through view-aware attention~\citep{zheng2023locally}, and autoregressive models enable interactive sketch-based mesh generation and editing~\citep{li2025meshpad}.
These methods target faithful reconstruction or plausible content creation, and are evaluated as such. Few address engineering directly, e.g., retrieving sketch-matched shapes for aerodynamic evaluation~\citep{Elrefaie25a} or inferring CAD part articulation from sketched arrows~\citep{yang2026sketch2arti}.

On the other hand, physics-aware 3D generation methods~\citep{Seiler26a,zhan2025physiopt} use the design input only to initialize the optimization, so the final geometry after optimization may reach the physical target while losing the intended styling.
The few methods that retain the sketch during optimization~\citep{Guillard21,You26a} tend to enforce strict fidelity to it: the same fixed sketch is injected at every iteration, e.g.\ as a conditioning input or through classifier-free guidance, instead of evolving together with the shape as the optimization proceeds.
This is limiting because it does not allow a designer to specify a specific style while allowing small deviations to increase performance.
\ours{} strikes a happy medium between these two extremes.

\section{AnchorGen}
\label{sec:method}
\begin{figure}[tb]
    \centering
    \includegraphics[width=\linewidth]{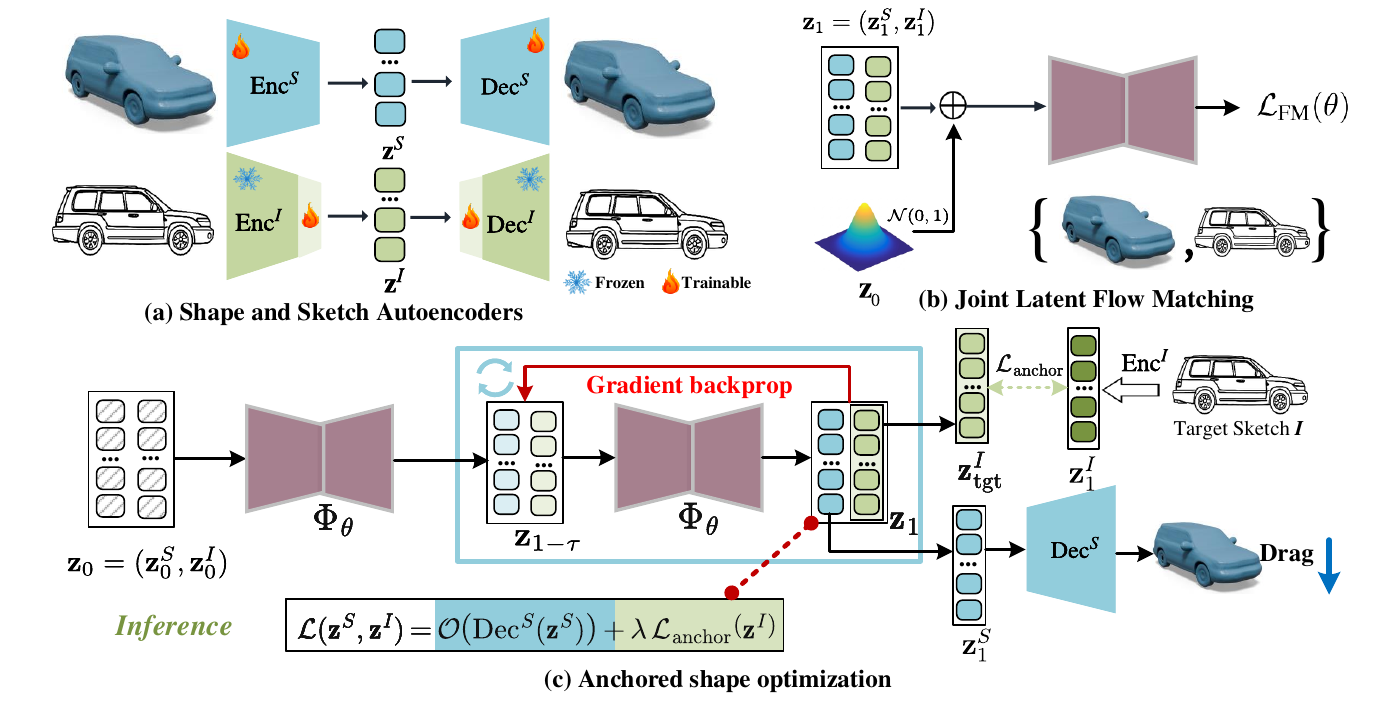}
    \vspace{-6mm}
    \caption{\small {\bf Framework.} (a) Two encoders, one for shape and one for sketches, are trained separately. (b) Given the encoders and their latent vectors, a flow matching algorithm is trained to produce them {\it jointly} from a single normal distribution. (c) At inference, a joint latent vector is generated, re-noised, and optimized to minimize the target physical objective while remaining close to the latent vector corresponding to the target sketch.
    }
    \label{fig:framework}
\end{figure}

Our goal is to generate a 3D shape $S$ that minimizes a numerical physical
objective such as aerodynamic drag, while preserving the original design intent
expressed as a target image sketch $I$. \ours{} encodes shapes and sketches into
latent vectors of the same size and trains a single flow matching model over
their concatenation, thereby modeling the \emph{joint} distribution of shapes and sketches. At inference time, it minimizes the objective \emph{through} that flow: The optimization variable is an intermediate state of the generative ODE, and the resulting pair it generates is scored by the objective on the decoded shape together with the distance from its sketch component to the target.
Every iterate is therefore an output of the joint prior, restricted to the shapes that matching the target sketch.
Fig.~\ref{fig:framework} summarizes these three stages, all of which operate in latent space, which keeps the dimensionality low and makes both the flow model and the optimization loop tractable.

\parag{Joint modeling of shape and sketch.}
We use an auto-encoding scheme to train two independent encoders to map a shape and a sketch to latents as $\mathbf{z}^{S}=\encd^{S}(S)$ and $\mathbf{z}^{I}=\encd^{I}(I)$. Both latent vectors are of the same dimension so that they can easily be concatenated.
The corresponding decoders map them back to a shape $\hat{S}=\decd^{S}(\mathbf{z}^{S})$ and to an image sketch $\hat{I}=\decd^{I}(\mathbf{z}^{I})$, such that $\hat{S}\approx S$ and $\hat{I}\approx I$, as shown in Fig.~\ref{fig:framework}(a).

We then use the standard Flow Matching (FM) approach~\citep{Lipman22,Liu23e} to learn a joint prior over $\mathbf{z}$ for all our training examples, as shown in Fig.~\ref{fig:framework}(b).
Each training pair is encoded once into the clean endpoint $\mathbf{z}_{1}=\left(\encd^{S}(S),\,\encd^{I}(I)\right)$, concatenated along the token axis, with modality embeddings added to identify the two kinds of tokens.
FM transports a simple prior $p_{0}=\mathcal{N}(\mathbf{0},\mathbf{I})$ at $t=0$ into the data distribution
$p_{1}$ at $t=1$ along the ODE whose vector field $\mathbf{v}(\mathbf{z}_{t})=\mathrm{d}\mathbf{z}_{t}/\mathrm{d}t$ is estimated from $\mathbf{z}_{t}$ and $t$ by a transformer $\mathbf{v}_{\theta}$.
Adopting the rectified-flow formulation, we sample $t\sim\mathcal{U}[0,1]$ and interpolate linearly between noise and data, so that the target conditional velocity reduces to a constant and the training objective becomes
\begin{equation}
\mathbf{z}_{t}=(1-t)\mathbf{z}_{0}+t\mathbf{z}_{1},
\qquad
\mathcal{L}_{\text{FM}}(\theta)=\mathbb{E}_{t,\mathbf{z}_{0},\mathbf{z}_{1}}
\left[\left\|\mathbf{v}_{\theta}(\mathbf{z}_{t},t)-(\mathbf{z}_{1}-\mathbf{z}_{0})\right\|^{2}\right].
\label{eq:fm}
\end{equation}
Training is fully unconditional, with both modalities noised and denoised jointly so that $\mathbf{v}_{\theta}$ learns their correspondence. We denote by $\Phi_{\theta}^{s\rightarrow u}$ the flow map integrating $\mathbf{v}_{\theta}$ from time $s$ to $u$.

\parag{Anchored shape optimization.}

Given a task-specific objective $\mathcal{O}:\mathcal{S}\rightarrow\mathbb{R}$, such as aerodynamic drag, and an initial pair $\mathbf{z}_{\mathrm{init}}=(\mathbf{z}^{S}_{\mathrm{init}},\mathbf{z}^{I}_{\mathrm{init}})$, our task can now be formulated as finding
\begin{equation}
\mathbf{z}_{*}=(\mathbf{z}^{S}_{*},\mathbf{z}^{I}_{*})
=\arg\min_{\mathbf{z}}\;\mathcal{O}\!\left(\decd^{S}(\mathbf{z}^{S})\right)
\quad\text{s.t.}\quad \decd^{I}(\mathbf{z}^{I})\approx I ,
\label{eq:anchor-opt-overview}
\end{equation}
and returning $S_*=\decd^{S}(\mathbf{z}^{S}_{*})$.
Since minimization is performed with our jointly trained flow model, the sketch latent $\mathbf{z}^{I}$ is part of the optimization rather than an external condition. Instead, it evolves together with the shape latent vector $\mathbf{z}^{S}$ while remaining anchored to the latent vector of the target sketch to ensure that the resulting sketch  resembles it.
To solve the problem of Eq.~\ref{eq:anchor-opt-overview}, we minimize
\begin{align}
\mathcal{L}( \mathbf{z}^{S},\mathbf{z}^{I}) &=  \mathcal{O}\!\left(\decd^{S}(\mathbf{z}^{S})\right)+\lambda\,\mathcal{L}_{\mathrm{anchor}}\!\left(\mathbf{z}^{I}\right) \; , \\
\mathcal{L}_{\mathrm{anchor}}\!\left(\mathbf{z}^{I}\right)
& =\frac{1}{N}\sum_{n=1}^{N}\left[\,1-
\frac{\left\langle \mathbf{z}^{I}_{n},\;\mathbf{z}^{I}_{\mathrm{tgt},n}\right\rangle}
{\left\|\mathbf{z}^{I}_{n}\right\|_{2}\,\left\|\mathbf{z}^{I}_{\mathrm{tgt},n}\right\|_{2}}\right] \; ,
\end{align}
where $\mathcal{L}_{\mathrm{anchor}}$ is the token-wise cosine distance to the target latent vector $\mathbf{z}^{I}_{\mathrm{tgt}}=\encd^{I}(I)$ over the $N$ sketch tokens, and $\lambda$ is a scalar controlling how far the sketch can deviate from its target.

\begin{wrapfigure}{R}{0.48\textwidth}
\vspace{-0.2\intextsep}
\begin{minipage}{\linewidth}
\begin{algorithm}[H]
\footnotesize
\setlength{\baselineskip}{9pt}
\raggedright
\algrenewcommand{\algorithmicindent}{1em}
\algrenewcommand{\algorithmiccomment}[1]{\ \ // #1}
\caption{Anchored Shape Optimization}
\label{alg:anchorgen}
\begin{algorithmic}[1]
\Require Target sketch $I$; optionally an initial shape $S_{\mathrm{init}}$ with its sketch $I_{\mathrm{init}}$
\Require Encoders $\encd^{S},\encd^{I}$, shape decoder $\decd^{S}$, flow $\mathbf{v}_{\theta}$ with flow map $\Phi_{\theta}$, objective $\mathcal{O}$
\Require Iterations $K$, step size $\eta$, anchor weight $\lambda$, inversion depth $\tau$
\If{$S_{\mathrm{init}}$ is given} \Comment{start from initial shape}
    \State $\mathbf{z}_{\mathrm{init}}\gets
           \left(\encd^{S}(S_{\mathrm{init}}),\,\encd^{I}(I_{\mathrm{init}})\right)$
\Else  \Comment{sample from scratch}
    \State $\mathbf{z}_{\mathrm{init}}\gets\Phi_{\theta}^{0\rightarrow1}(\mathbf{z}_{0}),
           \ \ \mathbf{z}_{0}\sim\mathcal{N}(\mathbf{0},\mathbf{I})$
\EndIf
\State $\mathbf{z}^{I}_{\mathrm{tgt}}\gets\encd^{I}(I)$
\State $\mathbf{z}_{1-\tau}\gets\Phi_{\theta}^{1\rightarrow1-\tau}(\mathbf{z}_{\mathrm{init}})$
       \Comment{invert to $t=1-\tau$}
\For{$k=0,\ldots,K-1$}
    \State $\left(\mathbf{z}^{S},\mathbf{z}^{I}\right)\gets
           \Phi_{\theta}^{1-\tau\rightarrow1}(\mathbf{z}_{1-\tau})$
           \Comment{back to $t=1$}
    \State $\mathcal{L}\gets
       \colorbox{lightblue}{$\mathcal{O}\!\left(\decd^{S}(\mathbf{z}^{S})\right)$}
       +\colorbox{lightgreen}{$\lambda\,\mathcal{L}_{\mathrm{anchor}}\!\left(\mathbf{z}^{I}\right)$}$
    \State $\mathbf{z}_{1-\tau}\gets\mathbf{z}_{1-\tau}-\eta\,\nabla_{\mathbf{z}_{1-\tau}}\mathcal{L}$
\EndFor
\State $\left(\mathbf{z}^{S}_{*},\mathbf{z}^{I}_{*}\right)\gets
       \Phi_{\theta}^{1-\tau\rightarrow1}(\mathbf{z}_{1-\tau})$
       \Comment{final pass}
\State \textbf{return} $S^{*}=\decd^{S}(\mathbf{z}^{S}_{*})$
\end{algorithmic}
\end{algorithm}
\end{minipage}
\vspace{-16mm}
\end{wrapfigure}

Crucially, we minimize $\mathcal{L}$ not with respect to $\mathbf{z}$ directly, but with respect to an intermediate state of the flow ( Alg.~\ref{alg:anchorgen}, Fig.~\ref{fig:framework}(c)).
We first obtain an initial pair $\mathbf{z}_{\mathrm{init}}$, either by encoding an existing design and its sketch, or by sampling noise and integrating the flow to $t=1$, and we encode the target sketch once into $\mathbf{z}^{I}_{\mathrm{tgt}}$.
We then invert $\mathbf{z}_{\mathrm{init}}$ back to an intermediate time $t=1-\tau$ and take the state $\mathbf{z}_{1-\tau}$ at that time as our optimization variable.
At each iteration, we integrate $\mathbf{z}_{1-\tau}$ forward to $t=1$ to recover a clean pair $(\mathbf{z}^{S},\mathbf{z}^{I})$, evaluate $\mathcal{L}$ on it, and backpropagate through the flow map to update $\mathbf{z}_{1-\tau}$.
A final integration and decoding of $\mathbf{z}^{S}_{*}$ yields the optimized shape $S^{*}$.

\parag{Why optimize through the flow.}

A more traditional approach would be to minimize
$\mathcal{O}\!\left(\decd^{S}(\mathbf{z}^{S})\right)$ with respect to $\mathbf{z}^{S}$ directly, while constraining the resulting shape $\decd^{S}(\mathbf{z}^{S})$ to match the target sketch.
Each gradient step then follows the objective alone, without any influence from the shape prior, and the iterate is projected back onto the manifold of valid shapes only afterwards, as in~\citep{Seiler26a}.
This decoupling has two drawbacks. The projection targets the manifold as a whole and cannot single out the region consistent with the design intent, so repeated corrections collapse towards the generic designs that dominate the prior.
Additionally, once the drift is large, recovering a valid, high-quality candidate strongly depends on the step size.
By making the flow part of the update itself, every proposed move is re-expressed by the \emph{joint prior} before it is scored, so that the iterate never leaves the manifold in the first place, and the anchor term makes the correction point to the slice of that manifold selected by the target sketch.

\parag{Implementation details.}
Shapes are encoded and decoded with the VecSet autoencoder~\citep{Zhang23d}, using the same architecture as \citet{Seiler26a}. For sketches, we freeze the Stable Diffusion autoencoder~\citep{Rombach22a} and train only a lightweight $1.8$M-parameter bottleneck that patchifies its $\mathbb{R}^{4\times32\times32}$ latent into $256$ tokens of $8$ channels, yielding $\mathbf{z}^{I}\in\mathbb{R}^{256\times8}$. The joint prior $\mathbf{v}_\theta$ is an $82$M-parameter transformer. We set $\tau=1$ on ShapeNet to accommodate its greater shape diversity and $\tau=0.4$ on DrivAerNet++. Further details are provided in Appendix~\ref{sec:implementation-details}, ablations on $\tau$ and model size in Appendix~\ref{sec:ablation}.

\section{Experiments}
\label{sec:experiments}

We evaluate the three modes of operation introduced in Sec.~\ref{sec:introduction} using a single trained model, using the setup of Sec.~\ref{sec:setup}. In Sec.~\ref{sec:anchored-optimization}, we demonstrate that anchoring to an existing sketch suppresses manifold drift and preserves design intent, as shown  in Fig.~\ref{fig:anchorgen-teaser}(a). In  Sec.~\ref{sec:customized-design}, we further show that a shape attribute can be added or changed by anchoring to an edited sketch, as shown in Fig.~\ref{fig:anchorgen-teaser}(b). Finally, in Sec.~\ref{sec:generation-from-scratch}, we demonstrate synthesis from scratch given only a sketch, as shown in  Fig.~\ref{fig:from_strach}. ShapeNet features a diverse inter-class design space and about 50\% of the baselines successfully manage a $30\%$ drag reduction. On DrivAerNet++, which is less diverse, about 50\% of the baselines deliver a $20\%$ drag reduction. Thus, these reduction numbers are the largest ones for which we can make a meaningful comparison and the ones we use in this section.  We provide results for lesser values of the drag reduction in Appendix~\ref{sec:customized-design-analysis}.

\subsection{Experimental Setup}
\label{sec:setup}

\parag{Datasets.}
We use two complementary vehicle datasets: ShapeNet~\citep{Chang15}, which emphasizes geometric diversity across car types, and DrivAerNet++~\citep{elrefaie2024drivaernet++}, which provides an explicit parameterized design space with high-fidelity aerodynamics.
On ShapeNet, we follow the setup of~\citep{Seiler26a}, including its surrogate model for drag evaluation. On DrivAerNet++, where every design comes with explicit length, width and height parameters and high-fidelity CFD annotations, we train the shape prior and the surrogate model on its training split.
For every shape, we render a normal map with a fixed orthographic camera and prompt Qwen~\citep{wu2025qwen}, which yields a sketch consistent with the silhouette, proportions and part semantics of the shape. Appendix~\ref{app:dataset} gives further details on data preprocessing and sketch generation.

\parag{Baselines.}
We compare against the two families of Sec.~\ref{sec:related}.
\emph{Prior-based} methods steer a shape-only prior at test time and differ in
how the physical objective is descended: gradient descent (GD)~\citep{zhan2025physiopt},
Flow Matching Guidance (FMG)~\citep{Ho22a}, D-Flow~\citep{BenHamu24},
ICTM~\citep{Zhang25e}, SGO~\citep{chen2026optimization}, and
FCSO~\citep{Seiler26a}. We evaluate them on both benchmarks.
\emph{Conditioning-based} methods instead learn a conditional distribution
$P(Y \mid X)$ of shapes given sketches, making the sketch an explicit input at
inference time, and differ in how the sketch injects a direction into the
optimization: through plain input conditioning~\citep{du2025sgcdiff}, or through
classifier-free guidance (CFG) with guidance-weight schedules~\citep{xi2024analysis}
or modified update rules~\citep{sadat2024eliminating,fan2025cfg,saini2026rectified,wang2026cfg}.
Since they require a sketch that encodes a design intent, we evaluate them on
DrivAerNet++ only. Appendix~\ref{sec:baseline_details} lists all variants and
implementation details.

\parag{Metrics.}
An optimized shape must stay physically plausible, so we evaluate results on two levels. First, we adopt the metrics and the experimental setup of~\citep{Seiler26a} to gauge whether an optimized shape remains in distribution. Second, for pairwise metrics, we compare each optimized shape to the shape used to produce the anchoring sketch in terms of F@$1\%$ and F@$0.5\%$, Chamfer distance (CD-L1), $95$th-percentile Hausdorff distance (HD95), normal consistency (Normal-C), volume ratio (Vol-Ratio), and the average controlled-dimension error (Avg.\ Dim-Err), which isolates a single controlled attribute such as vehicle length. All metrics are computed on the first output that attains the prescribed drag reduction. See Appendix~\ref{sec:metrics} for precise definitions.

\subsection{Anchoring to an Existing Sketch}
\label{sec:anchored-optimization}
 \begin{table}[t]
      \centering
      \caption{
    \textbf{Distributional Similarity.} Car drag optimization at $30\%$ drag
    reduction. Metrics are scaled:
    $\text{KD}_{\text{incep}}$ ($\times 10^3$),
    $\text{FD}_{\text{dino}}$ ($\times 10^{-1}$),
    $\text{FD}_{\text{point}}$ ($\times 10^2$), and
    $\text{KD}_{\text{point}}$ ($\times 10^5$).
}
      \label{tab:drag_distribution}
      \vspace{2mm}
      \small
      \setlength{\tabcolsep}{4pt}
      \begin{tabular*}{\textwidth}{@{\extracolsep{\fill}} l c c c c c c}
          \toprule
          \textbf{Method}
          & \textbf{FD}$_{\text{incep}}$ $\downarrow$
          & \textbf{KD}$_{\text{incep}}$ $\downarrow$
          & \textbf{FD}$_{\text{dino}}$ $\downarrow$
          & \textbf{KD}$_{\text{dino}}$ $\downarrow$
          & \textbf{FD}$_{\text{point}}$ $\downarrow$
          & \textbf{KD}$_{\text{point}}$ $\downarrow$ \\
          \midrule
          GD~\citeyearpar{zhan2025physiopt}
          & 132.82 & 106.26 & 97.97 & 75.63 & 37.50 & 123.20 \\
          FMG~\citeyearpar{Ho22a}
          & 100.20 & 67.02 & 77.05 & 56.52 & 25.00 & 73.00 \\
          D-Flow~\citeyearpar{BenHamu24}
          & 108.79 & 76.44 & 82.85 & 62.68 & 29.95 & 76.70 \\
          ICTM~\citeyearpar{Zhang25e}
          & 101.73 & 75.52 & 75.71 & 58.31 & 24.97 & 77.60 \\
          SGO~\citeyearpar{chen2026optimization}
          & 107.72 & 76.89 & 81.54 & 61.25 & 30.75 & 83.70 \\
          FCSO~\citeyearpar{Seiler26a}
          & 89.08 & 55.91 & 69.20 & 52.72 & 20.96 & 51.40 \\
          \midrule
          AnchorGen (Ours)
          & \textbf{77.74} & \textbf{44.82} & \textbf{58.56}
          & \textbf{44.23} & \textbf{15.75} & \textbf{33.07} \\
          \bottomrule
      \end{tabular*}
      \vspace{-2mm}
  \end{table}

\begin{figure}[t]
\centering
\includegraphics[width=\linewidth, trim=0 2mm 0 2mm, clip]{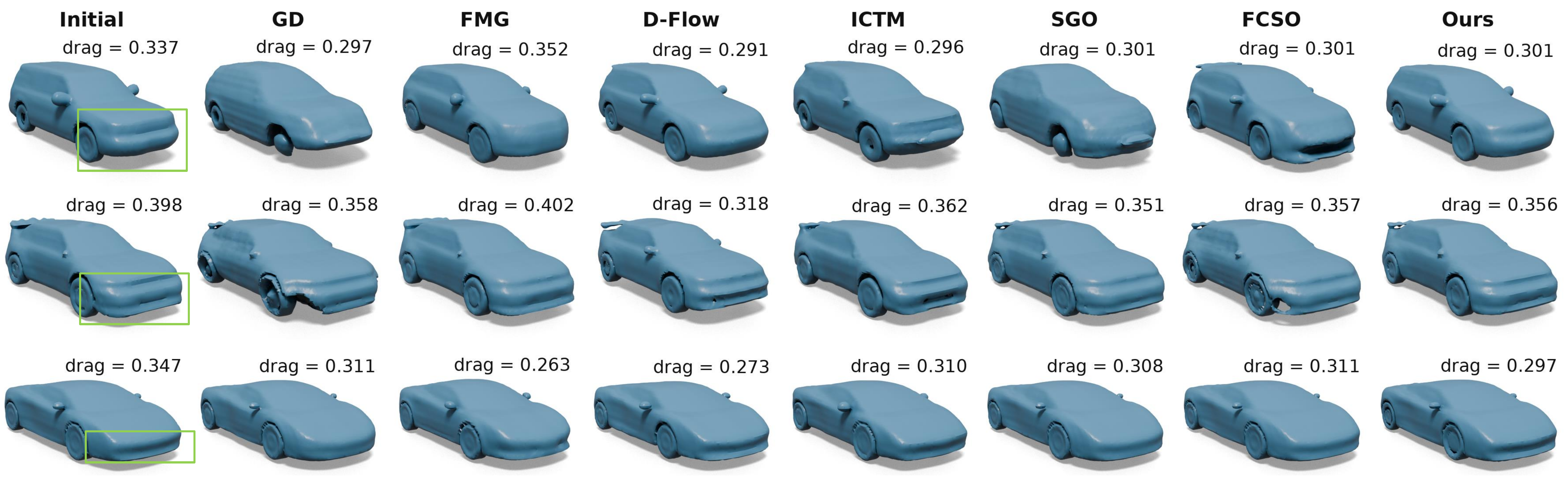}
\vspace{-6mm}
\caption{\textbf{Preservation of shape identity} at a $10\%$ drag reduction.
Green boxes mark identity features of the initial shape.}
\label{fig:shape_vis}
\vspace{-4mm}
\end{figure}

We explore two different ways to preserve style and avoid manifold drift when minimizing the drag of an existing vehicle.  One can either use the actual sketch of the original vehicle as the anchor or use the sketch of a lower-drag vehicle of the same type to guide the optimization in the desired direction. This mirrors design practice, where a designer knows, for instance, that moderately lowering the roof usually improves drag. We test this on ShapeNet, which is highly diverse, featuring all kinds of different cars ranging from compact sedans to trucks. As a result, optimization can easily alter the vehicle type or produce distorted shapes, making ShapeNet a sensitive probe for manifold drift.

\parag{Anchoring to the vehicle's own sketch.}

Fig.~\ref{fig:shape_vis} compares our results against the baselines at $10\%$ drag reduction. Green boxes mark examples of identity features of the initial shape, such as the front bumper. Since all baselines rely on a shape prior alone to correct the geometry, none of them reliably retains these features: GD and D-Flow smooth them out, ICTM and SGO distort them, and FCSO even grows parts the source never had, such as a rear wing. \ours{}, in contrast, retains them in all three examples. See Appendix~\ref{sec:shapenet_ablation} for quantitative results and ablations.

\parag{Anchoring to the sketch of a better-performing vehicle.}

Performance can be further improved when \ours{} is anchored to the sketch of a lower-drag vehicle, instead of the sketch of the original shape. This gives the optimization an explicit direction of improvement.
To select a suitable sketch for each one $100$ test cases, we collect from the training split the shapes whose drag is $30$--$40\%$ lower than that of the source as a candidate pool, from which we draw $36$ target sketches that stand in for the sketch a designer would produce from experience. See Appendix~\ref{par:sketch_matched} for details.  Since all sketches come from the training split, no data leakage is introduced; they only add an explicit direction for the optimization to follow, rather than the guidance of a shape prior alone. As shown in Table~\ref{tab:drag_distribution}, \ours{} improves all six metrics across all competitors.
Anchoring to a single target sketch rather than to the distribution as a whole also makes the optimization more reliable: FCSO relies on a shape prior alone to stay on the manifold, which tends to obstruct the descent, so that it reaches the 30\% reduction for only 88 of the 100 sources, whereas ours succeeds on 98.

\subsection{Anchoring to an Edited Sketch}
\label{sec:customized-design}

  \begin{table}[t]
  \centering
\caption{\textbf{Pairwise similarity for dimension edits on DrivAerNet++.}
Similarity to the target shape over $600$ source--target pairs at a $20\%$ drag
reduction, averaged over the three views.
\emph{Top:} prior-based methods, which steer a shape-only prior at test time.
\emph{Middle:} conditioning-based methods, which train a sketch-conditioned
prior and inject the sketch via input conditioning or classifier-free guidance.
\emph{Bottom:} \ours{}, which trains a single prior over shape-sketch pairs.
Metrics are scaled by $\times 100$; NaN marks an optimization that became
unstable and collapsed.}
  \label{tab:total_metrics_scaled}
  \setlength{\tabcolsep}{4pt}
  \resizebox{\textwidth}{!}{%
  \begin{tabular}{l|ccccccc}
  \toprule
  \textbf{Method} &
  \textbf{F@1\%} $\uparrow$ &
  \textbf{F@0.5\%} $\uparrow$ &
  \textbf{CD-L1} $\downarrow$ &
  \textbf{Vol-Ratio} $\uparrow$ &
  \textbf{Normal-C} $\uparrow$ &
  \textbf{HD95} $\downarrow$ &
  \textbf{Avg. Dim-Err} $\downarrow$ \\
  \midrule
  GD~\citeyearpar{zhan2025physiopt}
  & 73.30 & 47.22 & 1.672 & 91.88 & 89.44 & 2.576 & 4.69 \\

  FMG$^\dagger$~\citeyearpar{Ho22a}
  & NaN & NaN & NaN & NaN & NaN & NaN & NaN \\

  D-Flow~\citeyearpar{BenHamu24}
  & 65.91 & 44.64 & 1.923 & 88.06 & 87.81 & 2.821 & 4.18 \\

  ICTM$^\dagger$~\citeyearpar{Zhang25e}
  & NaN & NaN & NaN & NaN & NaN & NaN & NaN \\

  SGO$^\dagger$~\citeyearpar{chen2026optimization}
  & NaN & NaN & NaN & NaN & NaN & NaN & NaN \\

  FCSO~\citeyearpar{Seiler26a}
  & 73.44 & 48.24 & 1.661 & 92.04 & 89.44 & 2.587 & 4.37 \\
  \midrule
  Input Cond.~\citeyearpar{du2025sgcdiff}
  & 71.67 & 45.68 & 1.713 & 90.73 & 89.09 & 2.577 & 4.00 \\

  Constant~\citeyearpar{xi2024analysis}
  & 72.80 & 46.62 & 1.691 & 90.46 & 89.23 & 2.560 & 3.91 \\

  Linear Schedule~\citeyearpar{xi2024analysis}
  & 73.58 & 47.48 & 1.665 & 90.40 & 89.42 & 2.526 & 3.67 \\

  Exponential Schedule~\citeyearpar{xi2024analysis}
  & 73.20 & 46.97 & 1.679 & 90.43 & 89.32 & 2.543 & 3.84 \\

  APG~\citeyearpar{sadat2024eliminating}
  & 73.74 & 47.68 & 1.659 & 90.44 & 89.46 & 2.518 & 3.64 \\

  CFG-Zero$^\star$~\citeyearpar{fan2025cfg}
  & 72.77 & 46.57 & 1.692 & 90.52 & 89.21 & 2.559 & 3.87 \\

  Rectified-CFG++~\citeyearpar{saini2026rectified}
  & 72.36 & 46.14 & 1.706 & 90.47 & 89.12 & 2.579 & 4.02 \\

  SMC~\citeyearpar{wang2026cfg}
  & 69.67 & 44.20 & 1.795 & 90.42 & 88.47 & 2.691 & 4.49 \\
  \midrule
  AnchorGen (Ours)
  & \textbf{79.68}
  & \textbf{55.60}
  & \textbf{1.426}
  & \textbf{92.62}
  & \textbf{91.01}
  & \textbf{2.392}
  & \textbf{2.67} \\
  \bottomrule
  \end{tabular}%
  }
  \vspace{-2mm}
  \end{table}

To demonstrate that \ours{} can be used to re-design an existing vehicle by editing its sketch, we use DrivAerNet++~\citep{elrefaie2024drivaernet++} whose designs are all obtained by morphing a small set of DrivAer baselines~\citep{Heft12} within a bounded parametric design space. Each sample comes with explicit dimension parameters such as length, width and roof height, which lets us express design intent as a quantified target an engineer would actually specify.

\parag{Dimensional edits.}
\label{par:quantified_customized_design}

We randomly select $100$ source vehicles from the test split and pair each with six targets, obtained by stretching or compressing one of length (side view), width (front view) or height (rear view), yielding $600$ source-target pairs on which we measure the change in the corresponding dimension. Table~\ref{tab:total_metrics_scaled} reports results for both baseline families at a $20\%$ drag reduction, averaged over the three views. \ours{} is best on every metric.
On Avg.\ Dim-Err in particular, it outperforms the conditioning-based variants, which in turn outperform the prior-based methods that supply no explicit direction.
Several baselines become unstable under this finer-grained optimization: ICTM and FMG, which steer only toward a nominal drag value, rarely descend all the way to a $20\%$ reduction, while SGO enforces no strict manifold projection and is prone to shape collapse. Per-view results and metrics tracked along the optimization trajectory are given in Appendix~\ref{sec:customized-design-analysis}.

\parag{Free-form edits.}

A sketch can also express things fixed set of scalar parameters cannot. As shown in Fig.~\ref{fig:customized_design}, we add a rear wing to a ShapeNet car without letting drag blow up, by drawing the wing in red on top of the sketch rendered from the source. This stroke expresses design intent rather than an aerodynamically optimal configuration, yet our method resolves the conflict on its own: the final shape and its sketch converge to a configuration that satisfies the design requirement while still reducing drag by $9.83\%$.

\begin{figure}[tb]
	\centering
	\includegraphics[width=1\linewidth]{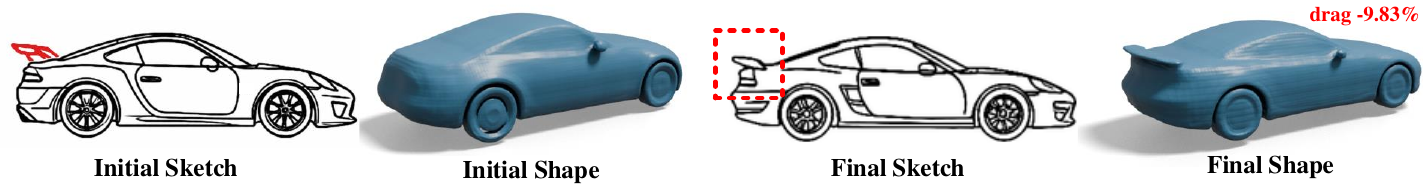}
    \vspace{-7mm}
	\caption{\textbf{Customized Design under a Drag Objective.} Starting from a base vehicle shape with a user-added rear spoiler, we show how the shape evolves under the drag optimization objective while the specified modification is preserved.
	}
	\label{fig:customized_design}
    \vspace{-6mm}
\end{figure}

\subsection{Synthesis from a sketch}
\label{sec:generation-from-scratch}
\begin{figure}[tb]
	\centering
	\includegraphics[width=\linewidth]{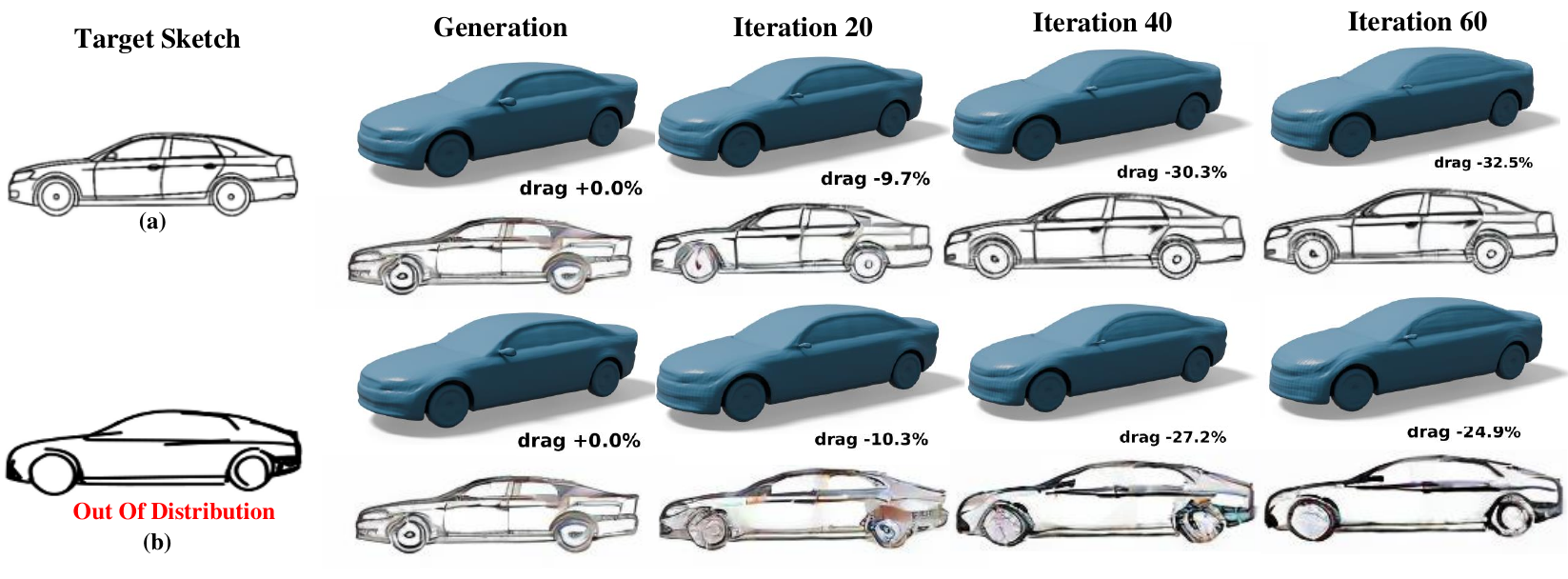}
    \vspace{-7mm}
    \caption{\textbf{Generation from scratch.} Generation starts from
	noise, with the shape and decoded sketch shown over $60$ sketch-anchored optimization iterations. (a) Training-style sketches. (b) Simple line
	drawings in a zero-shot setting.}
	\label{fig:from_strach}
    \vspace{-4mm}
\end{figure}

\ours{} is not restricted to refining an existing shape. In Fig.~\ref{fig:from_strach}(a), we sample a shape from the unconditional prior as the starting point and let the target sketch anchor the drag optimization. Along the trajectory, the predicted drag decreases steadily while the geometry moves toward the intended design: the cabin gradually merges into a fastback profile that runs continuously into the rear end, matching the drawing.
To test out-of-distribution sketches, we further use the CLIPasso~\citep{vinker2022clipasso} sketches released with the DrivAerNet++ sketch extension, which abstract a car into a few contour strokes, a style far sparser than our training sketches and never seen during training. As shown in Fig.~\ref{fig:from_strach}(b), the model adapts to it on its own: the decoded sketch progressively takes on the sparse, few-stroke appearance of the guidance drawing while the shape still satisfies the design intent. This shows that the sketch anchor transfers beyond the exact style used for training. We further evaluate this setting \emph{quantitatively} on $10$ random shapes, using their CLIPasso sketches as targets and running both \ours{} and PhysGen~\citep{You26a} for $60$ cycles, with drag evaluated by OpenFOAM~\citep{jasak2007openfoam} (Table~\ref{tab:shape_metrics_side_physgen}). PhysGen attains slightly higher geometric similarity, likely owing to its larger pretrained model, whereas \ours{} is trained from scratch. Nevertheless, although these sketches are out of distribution for \ours{} but not for PhysGen, \ours{} achieves better length control and lower drag with $138$M vs.\ $400$M parameters, and runs about $12\times$ faster ($\sim1.5$ vs.\ $\sim18.5$\,min per optimization on one A100). See Appendix~\ref{sec:generation_from_scratch} for details.

\begin{table}[t]
  \centering
  \caption{\textbf{Generation from scratch, anchored to a given target sketch.}
  }
  \label{tab:shape_metrics_side_physgen}
  \setlength{\tabcolsep}{4pt}
  \resizebox{\textwidth}{!}{%
  \begin{tabular}{l|ccccccc|c}
  \toprule
  \textbf{Method}
  & \textbf{F@1\%} $\uparrow$
  & \textbf{F@0.5\%} $\uparrow$
  & \textbf{CD-L1} $\downarrow$
  & \textbf{Vol-Ratio}$^{*}$
  & \textbf{Normal-C} $\uparrow$
  & \textbf{HD95} $\downarrow$
  & \textbf{L-Err} $\downarrow$
  & \textbf{Drag (CFD)} $\downarrow$ \\
  \midrule
  PhysGen~\citeyearpar{You26a}
  & \textbf{64.66}
  & \textbf{42.56}
  & \textbf{1.85}
  & \textbf{104.12}
  & \textbf{89.55}
  & \textbf{2.46}
  & 3.21
  & 0.2505 \\
  AnchorGen (Ours)
  & 63.50
  & 40.31
  & 1.98
  & 89.26
  & 85.61
  & 2.54
  & \textbf{2.43}
  & \textbf{0.1894} \\
  \bottomrule
  \end{tabular}%
  }
  \vspace{-4mm}
\end{table}

\vspace{-2mm}
\subsection{Analysis}
\label{sec:analysis}
  \vspace{-2mm}
Additional experiments in the Appendix further demonstrate that the improvements stem from the direction supplied by the anchor rather than from the prior alone. The sketch-to-drag ratio $\lambda$ trades adherence to the target sketch against drag reduction: a larger weight brings the result closer to an exogenous target, whereas anchoring too strongly to the source's own sketch confines the search and slows the drag descent (Appendix~\ref{sec:shapenet_ablation}). Edits obtained with shape-prior-only correction collapse towards the high-density region of the data, whereas ours spread along the directions prescribed by the target sketches (Appendix~\ref{par:diversity_of_controlled_edits}). \ours{} scales with parameter count, yet capacity alone does not close the gap: our smallest model, with $14$M parameters, already outperforms an $82$M-parameter FCSO on all seven metrics of Table~\ref{tab:total_metrics_scaled} (Appendix~\ref{sec:ablation}).

 \vspace{-2mm}
\section{Conclusion}

We introduced \ours{}, which turns the design sketch from an initialization into an anchor that persists throughout physics-driven shape optimization. A single unconditional flow over shape-sketch pairs, constrained only at inference, directs the corrections that keep a design valid toward the shapes the sketch specifies rather than the high-density region of the data. Thus, one model supports optimization anchored to an existing or edited sketch, and synthesis from a sketch alone. While we demonstrated the approach on drag minimization of cars, the framework is generic, and we will extend it to further objectives and shape categories, such as chair volume optimization.

\subsection*{AI use statement}
In this work, we used generative AI tools for generating synthetic data sets, implementing methods, assisting with translation, and cleaning and reformatting datasets. We have not used generative AI tools to help develop theoretical models or conceptual frameworks, formulate mathematical claims, propose or refine hypotheses, design or provide feedback on research methodology or experiments, support qualitative and thematic data analysis, or interpret results; providing critical ingredients for proving mathematical claims and assisting in the writing of proofs are not applicable to this work. Additionally, we used generative AI tools for creating or modifying scientific figures or images, and for creating or editing software code. We have reviewed all AI-assisted work. We checked the quality and style of the AI-generated sketches synthesized from shapes, multiple co-authors checked the text translations provided by LLMs, and we verified the AI-assisted code for the implementation of parts of our method and the 3D mesh visualization pipeline. We take responsibility for the final content of this work, including text, claims or artifacts produced with the aid of generative AI.

\subsection*{Ethics statement}

All experiments in this study were conducted on publicly available datasets. The only additional data we created are sketches synthesized from the shapes in these datasets; no human subjects, personal information, or privacy-sensitive data were involved. We acknowledge that, like other generative 3D models, our method could be misused to create deceptive content, for example, fabricated shapes spread as misinformation on social media, or falsified evidence of vehicle damage submitted to insurance companies to commit fraud.

\subsection*{Reproducibility statement}
In the appendix, we provide detailed descriptions of the metric computation
(Appendix~\ref{sec:metrics}); the dataset preprocessing, sketch generation, and
target matching (Appendix~\ref{app:dataset}); the implementation of the compared
baselines (Appendix~\ref{sec:baseline_details}); the training procedures of the
respective models (Appendices~\ref{sec:surrogate}
and~\ref{sec:implementation-details}); the ablation studies and hyperparameter
selection (Appendices~\ref{sec:shapenet_ablation} and~\ref{sec:ablation}); and
the CFD simulation setup (Appendix~\ref{sec:generation_from_scratch}). Upon
acceptance, we will make the preprocessed datasets and the corresponding code
publicly available.

\bibliographystyle{preprint}
\bibliography{references}

\clearpage
\appendix

\addtocontents{toc}{\protect\setcounter{tocdepth}{2}}%
\begingroup
  \renewcommand{\contentsname}{Contents of the Appendix}%
  \setcounter{tocdepth}{-1}%
  \tableofcontents
\endgroup
\newpage

\section{Metrics}
\label{sec:metrics}

\parag{Shape metrics.}
 Each optimized shape is compared against its paired target in the common
  decoder coordinate frame, without rigid alignment, ICP, or per-object
  rescaling. From each mesh extracted by evaluating the decoded signed distance
  field on a regular grid and applying Marching Cubes at the zero level set,
  we draw $60{,}000$ surface samples with fixed seeds, each carrying the face
  normal of its source triangle, and match every sample to its nearest neighbor
  on the other surface. We write
  $d^{p\rightarrow t}$ for the distances from predicted samples to the target
  surface and $d^{t\rightarrow p}$ for the reverse direction. All distances are
  normalized by $s$, the diagonal of the target's axis-aligned bounding box
  (AABB); normalizing by the target rather than by each mesh individually keeps
  scale discrepancies between prediction and target visible. CD-L1 sums the two directional means,
\begin{equation}
\mathrm{CD\text{-}L1}=\frac{1}{s}\left[
\operatorname{mean}\big(d^{p\rightarrow t}\big)+
\operatorname{mean}\big(d^{t\rightarrow p}\big)\right],
\end{equation}
where L1 denotes unsquared distances, not a coordinate-wise Manhattan metric.
F@$1\%$ and F@$0.5\%$ are the harmonic means of precision and recall, counting
a sample as matched when its nearest-neighbor distance falls below $0.01s$ and
$0.005s$ respectively; precision measures how much of the predicted surface
lies near the target, recall how much of the target is covered. HD95 is the
larger of the two directional $95$th-percentile distances, again divided by
$s$---a robust variant of the Hausdorff distance that tolerates a few outlying
samples. Normal-C averages, over both directions, the absolute dot product
between the normals of matched samples. Vol-Ratio divides the absolute
predicted mesh volume by the absolute target volume and is optimal at $1$.

\parag{Controlled-dimension error.}
Under the DrivAerNet++ Z-up convention, the $x$-, $y$- and $z$-extents of a
mesh AABB correspond to vehicle length, width and height. Every pair is scored
only on its designated controlled axis---length for side-view, width for
front-view, height for rear-view pairs. Writing $D_i^{\mathrm{pred}}$ and
$D_i^{\mathrm{target}}$ for the extents of the $i$-th pair along that axis,
\begin{equation}
\mathrm{Avg.\ Dim\text{-}Err}=\frac{1}{N}\sum_{i=1}^{N}
\frac{\left\vert D_i^{\mathrm{pred}}-D_i^{\mathrm{target}}\right\vert}
{D_i^{\mathrm{target}}},
\end{equation}
the mean relative absolute error on the controlled dimension over the $N$
evaluation pairs. It does not penalize drift along the two remaining axes.

\section{Dataset}
\label{app:dataset}

\paragraph{Data Preprocessing.}
\label{sec:data_preprocessing}

\textit{ShapeNet:} We use the car category of ShapeNet~\citep{Chang15}, which
spans a wide range of body styles and topologies but provides no design
parameterization, and therefore probes how well a shape prior generalizes when
the manifold of valid shapes is broad and unstructured. We adopt the same data
and split as FCSO~\citep{Seiler26a}: meshes are made watertight, aligned and
normalized to $[-1,1]^3$, and the corresponding drag coefficients are obtained
with OpenFOAM~\citep{jasak2007openfoam}. This yields $836$ shapes for training the shape
prior and the drag-conditioned flow matching model, on top of which we report
the optimization process on $100$ test cases.

\textit{DrivAerNet++}~\citep{elrefaie2024drivaernet++} is built by morphing a
family of parametric DrivAer models, so every design is fully specified by a set
of design parameters covering overall dimensions such as length, width and roof
height as well as discrete configuration choices such as the rear-end type and
the underbody treatment. We discard the detailed-underbody designs, whose
variation lies in the chassis structure rather than in the exterior shape we
optimize, and retain $4{,}165$ designs, partitioned into $3{,}051$ training,
$555$ validation and $559$ test samples. The shape prior and the surrogate model
are trained on the training split.

\paragraph{Sketch generation.}
All sketches in this work are synthetic and derived from 3D meshes through a
fixed pipeline. We place each mesh under a fixed orthographic camera
convention and render a $1024\times768$ normal map, using consistently
oriented cell normals transformed into camera coordinates, with each component
mapped linearly from $[-1,1]$ to RGB. Depending on the experiment we render a
single canonical viewpoint or several complementary ones; only the normal map
is passed downstream. Each normal map is converted into a black-and-white line drawing with
\texttt{Qwen/Qwen-Image-Edit-2511}\footnote{Model weights: \url{https://huggingface.co/Qwen/Qwen-Image-Edit-2511}.} ~\citep{wu2025qwen} in image-edit mode, the normal render being
the only image input. We use seed $7$, $30$ inference steps,
\texttt{true\_cfg\_scale}$=3.0$, \texttt{guidance\_scale}$=1.0$, a working
canvas of long edge $512$, and an input content margin of $0.12$. The prompt,
reproduced below, instructs the model to trace only the visible parts and
forbids adding, removing, relocating or restyling any component.

Writing $g$ for the grayscale value of the generated image, we binarize with
$\alpha=\mathrm{clip}\!\left((T-g)/r,\,0,\,1\right)$ and
$I_{\mathrm{line}}=255\,(1-\alpha)$, using $T=240$ and $r=30$ to preserve
anti-aliasing. We then crop the tight bounding box of all non-white content
and rescale it isotropically onto a $256\times256$ canvas with a $7\%$ margin. This per-image fitting normalizes the pixel scale of each vehicle, so image
space alone cannot express absolute dimensions. Where a sketch must control
them, we additionally register the line drawing to the projection of its
paired 3D bounding box: the world cube $[-1,1]^{3}$ is mapped at $119.5$
pixels per world unit about the canvas center, and the ink bounding box
$(l_s,t_s,r_s,b_s)$ is rescaled onto the projected target box
$(l_t,t_t,r_t,b_t)$ with $s_x=(r_t-l_t)/(r_s-l_s)$ and
$s_y=(b_t-t_t)/(b_s-t_s)$. Since $s_x$ and $s_y$ are independent, the
registration is axis-aligned affine and may be anisotropic. Sketches that only
convey silhouette, proportion and part semantics are left unregistered. The
prompt passed to the image-edit model is:

\begin{quote}\small\itshape
Redraw this car normal map as a clean black-and-white line drawing. Reproduce the
car exactly as it appears in the input: trace only the outlines, panel lines,
window and light edges, wheels, and every other part and protrusion that is
actually visible in the input. Keep the silhouette, proportions, viewpoint, scale,
framing, and the complete part set identical: each visible part stays in its
original position, size, and shape. Do not add, remove, relocate, reinterpret,
redesign, or restyle any part, and do not introduce any part that is not present
in the input. Black outlines on a plain white background, no shading and no fill.
\end{quote}

\paragraph{Target matching.}
\label{par:sketch_matched}

\textit{ShapeNet.}
For \emph{Anchoring to an Existing Sketch} we take the $100$ ShapeNet
sources of the FCSO protocol and retrieve one non-self target for each from
the shape reference pool, using the pressure-integrated drag predicted by the
GCNN surrogate on the original meshes. For a source $i$ and candidate $j$ we
write the relative drag reduction as
$r_{i\rightarrow j}=(\widehat C_D^{i}-\widehat C_D^{j})/\widehat C_D^{i}$ and
call a match strict when $0.30\leq r_{i\rightarrow j}\leq0.40$. Eligible
candidates are ranked lexicographically by the cosine distance between
side-view sketch embeddings, the $\ell_{2}$ distance between standardized
five-dimensional shape descriptors (three bounding-box extents, log surface
area, log absolute volume), and proximity to a nominal $30\%$ reduction. When
the strict window is empty we fall back to a $0.05$ band around the best
attainable reduction or the nearest window boundary, with the same ranking.
This yields $87$ strict and $13$ fallback matches, hence $100$ pairs with no
self matches and $36$ unique target UIDs. The optimizer receives the
side-view line drawing of the mapped target as its anchor; the target mesh,
SDF and shape latent are withheld.

\textit{DrivAerNet++.}
For \emph{Anchoring to an Edited Sketch} we build the pool from the $4{,}165$
DrivAerNet++ vehicles that have both the official design attributes and
complete registered front, side and rear sketches. We standardize
\texttt{A\_Car\_Length}, \texttt{A\_Car\_Width} and
\texttt{A\_Car\_Roof\_Height} over this pool. Of the vehicles, $274$ admit all six requested edits; from these we select $100$
sources by $k$-means in the standardized attribute space, preferring within
each cluster a source with large feasible edits and tight held-attribute
tolerances. Each source is paired with six non-self targets: longer and
shorter in the side view, wider and narrower in the front view, taller and
lower in the rear view. For each direction the matcher first constrains the
two held attributes and then maximizes the signed change of the controlled
one, trying the thresholds
$(\delta_{\mathrm{held}},\delta_{\mathrm{primary}})\in\{(0.10,1.20),
(0.15,1.00),(0.20,0.90),(0.25,0.80),(0.30,0.70),(0.40,0.60)\}$ in
standardized units from strict to loose; both held changes must stay below
$\delta_{\mathrm{held}}$ and the primary change must have the requested sign
and magnitude at least $\delta_{\mathrm{primary}}$. The resulting
$100\times6=600$ assignments comprise $433$ training-target and $167$
holdout-target rows. For every pair we reuse the target UID's existing
registered line drawing from the appropriate view. At test time the optimizer
sees only the target sketch, although that sketch was rendered from the target
geometry during benchmark construction.

\section{Baselines}
\label{sec:baseline_details}

\parag{Prior-injection baselines.}
This family of baselines compares the mechanisms by which a shape prior supplies
a direction to the optimization, with particular attention to controllability:
the sketch strength that each mechanism can actually attain in practice. All
variants use the same frozen encoders and the same target sketch, and differ
only in how that sketch reaches the velocity field.

\parag{Input conditioning.}
The simplest choice is to fix the conditioning signal at training time rather
than to intervene at inference. The conditional flow diffuses only the shape
latent and concatenates the clean sketch latent along the channel dimension, so
the model directly learns $p(z_{\mathrm{shape}} \mid
z_{\mathrm{sketch}})$. With noise at $t=0$ and
data at $t=1$, we set $x_t=(1-t)\epsilon+tz$ and train with
$\mathbb{E}\Vert v_\theta(x_t,t,c)-(z-\epsilon)\Vert_2^2$. Side, front, and rear
views use three separately trained models sharing the same backbone and
optimization setting. Because the condition is baked into the learned field, its
influence cannot be adjusted after training; no condition dropout is used here.

\parag{Classifier-free guidance.}
The remaining baselines move the intervention to inference time. An otherwise
identical model is retrained with the entire sketch latent zeroed with
probability $0.1$, which lets a single network produce both a conditional
velocity $v_c=v_\theta(x,t,c)$ and an unconditional one $v_u=v_\theta(x,t,0)$;
their difference then becomes an explicit, tunable direction. All variants below
reuse these same checkpoints and none trains a dedicated network, so they differ
purely in how the two branches are combined.

\begin{itemize}
    \item \textit{Constant, Linear, and Exponential
    schedules}~\citep{xi2024analysis}. These vary \emph{when} the guidance acts
    rather than what it points to, isolating the temporal weighting of one and
    the same CFG residual. All three apply $v=v_u+\omega(t)(v_c-v_u)$ with
    $\omega=s$, $\omega(t)=2ts$, and $\omega(t)=se^{t}/(e-1)$ respectively; the
    constant schedule gives a time-homogeneous reference, the linear one
    strengthens the condition as the clean endpoint is approached, and the
    exponential one redistributes weight nonlinearly through a convex time
    profile. We use $s=2$ throughout.

    \item \textit{APG}~\citep{sadat2024eliminating}. Rather than rescaling the
    residual as a whole, APG filters its geometry, controlling the magnitude and
    the direction of the update separately. Clean-endpoint estimates
    $\hat{x}_{1,\bullet}=x+(1-t)v_\bullet$ are formed from both branches; their
    difference is accumulated with momentum $\beta=0.5$, clipped at norm $5$,
    and reduced to the component orthogonal to $\hat{x}_{1,u}$ ($\eta=0$) before
    being converted back to a velocity.

    \item \textit{Rectified-CFG++}~\citep{saini2026rectified}. Its distinguishing
    feature is a half-step look-ahead: the correction is recomputed at a locally
    advanced rectified-flow state instead of taken from the branch difference at
    the current state. The conditional--unconditional correction is evaluated at
    an Euler midpoint with strength $\alpha(t)=\lambda_{\max}t^{\gamma}$, where
    $\lambda_{\max}=\gamma=1$, so its strength is governed entirely by these two
    parameters rather than by the generic guidance scale.

    \item \textit{CFG-Zero}$^\star$~\citep{fan2025cfg}. Instead of a prescribed
    schedule, the coefficient here is state-dependent, calibrated to the
    instantaneous alignment between the two branches. The unconditional branch is
    rescaled by $s^{*}=\langle v_c,v_u\rangle/(\Vert v_u\Vert_2^2+\epsilon)$,
    giving $v=(1-s)s^{*}v_u+sv_c$ with $s=2$, and the first step of each forward
    trajectory returns zero velocity.

    \item \textit{SMC}~\citep{wang2026cfg}. Guidance is cast as state feedback
    that depends not only on the current branch gap but also on how it changed
    since the previous ODE step, which sets SMC apart from the memoryless
    algebraic extrapolation of standard CFG. Treating $e_t=v_c-v_u$ as a control
    error, the sliding variable $q_t=(e_t-e_{t-1})+\lambda e_{t-1}$ yields the
    bounded, element-wise correction $v=v_u-sk\,q_t/(\vert q_t\vert+\epsilon)$,
    with $s=2$, $\lambda=5$, and $k=0.1$.
\end{itemize}

\section{Surrogate Model}
\label{sec:surrogate}

\parag{DrivAerNet++.} We use the cleaned version of the DrivAerNet++
dataset~\citep{elrefaie2024drivaernet++} and train a Transolver~\citep{Wu24a}
surrogate model for surface pressure prediction on its smooth-underbody subset. The
surrogate is trained on the 3{,}051 smooth-underbody vehicles of the corresponding training
split for 100 epochs. At every pass over the dataset, we sample 32{,}768 surface points per
vehicle without replacement. We minimize the mean squared error between the predicted
pressure and the normalized per-point pressure using Adam ($\texttt{eps}=1\mathrm{e}{-8}$,
weight decay $0$) with a two-phase cosine OneCycle schedule and an initial learning rate of
$4\mathrm{e}{-5}$. At inference time, we first transform the decoded vehicle mesh into the
DrivAerNet coordinate frame before querying the surrogate.

\section{Implementation details}
\label{sec:implementation-details}

\subsection{Autoencoder model}
\label{sec:autoencoder}
\paragraph{Shape branch.}
We adopt a VecSet-style neural field autoencoder~\citep{Zhang23d}.
The encoder $\encd^{S}$ takes a surface point cloud
$P\in\mathbb{R}^{2048\times3}$, selects $256$ anchor points by farthest point
sampling, and lets the corresponding tokens cross-attend to the full point set
at width $512$, yielding a set-structured latent
$\mathbf{z}^{S}\in\mathbb{R}^{256\times8}$. The decoder $\decd^{S}$ lifts
$\mathbf{z}^{S}$ from $8$ to $512$ channels, refines it with $12$
self-attention layers, and cross-attends it with arbitrary query coordinates to
produce a continuous signed distance field $f(\mathbf{p}\mid\mathbf{z}^{S})$; an
explicit mesh is recovered by marching cubes on a $256^{3}$ grid over
$[-1,1]^{3}$. We reuse the released checkpoint of \citet{Seiler26a} without
retraining.

\paragraph{Sketch branch.}
We keep a pre-trained autoencoder frozen and learn only a thin bottleneck that
reshapes its latent to the same token count and width as $\mathbf{z}^{S}$.
Concretely, $\encd^{I}$ composes the encoder $\mathcal{V}_{\mathrm{enc}}$ of the
KL-$f8$ autoencoder of Stable Diffusion~\citep{Rombach22a} with a learned
bottleneck $E_{b}$,
\begin{equation}
I\in\mathbb{R}^{3\times256\times256}
\;\xrightarrow{\;\mathcal{V}_{\mathrm{enc}}\;}\;\mathbb{R}^{4\times32\times32}
\;\xrightarrow{\;2\times2\text{ patchify}\;}\;\mathbb{R}^{256\times16}
\;\xrightarrow{\;E_{b}\;}\;\mathbf{z}^{I}\in\mathbb{R}^{256\times8},
\label{eq:sketch-encoder}
\end{equation}
and $\decd^{I}$ inverts it exactly: $D_{b}:8\!\rightarrow\!16$, unpatchify to
$\mathbb{R}^{4\times32\times32}$, then the frozen decoder
$\mathcal{V}_{\mathrm{dec}}$. The $2\times2$ patchification brings
$\mathbf{z}^{I}$ to the same token count as $\mathbf{z}^{S}$, leaving only a
channel-wise reduction $16\!\rightarrow\!8$ to be learned.

\paragraph{Training details.}
Only the bottleneck $\{E_{b},D_{b}\}$ is trained. It is a three-layer
convolutional residual block of width $128$ ($1.8$M parameters, against
$83.7$M frozen autoencoder parameters), optimized for $600$ epochs with Adam,
batch size $256$, and a learning rate decayed from $10^{-3}$ to
$5\!\times\!10^{-5}$ on a cosine schedule. The loss is a mean-squared error on
per-channel standardized latents, with the training-set statistics stored and
reused at inference so that $\encd^{I}$ and $\decd^{I}$ remain exact inverses.
\subsection{Gradient Combination}

The objective $\mathcal{O}$ and the anchor loss
$\mathcal{L}_{\mathrm{anchor}}$ frequently conflict, and naively summing
their gradients lets whichever term happens to have the larger magnitude
dominate the update. We therefore follow the multi-objective optimization
literature~\citep{desideri2012multiple,sener2018multi} and combine the two at the level of
gradients rather than losses. The objective gradient is first projected onto
the shape-token block, since $\mathcal{O}$ is defined only through
$\decd^{S}$ and carries no meaningful signal on the sketch tokens. Both
gradients are then normalized to unit length, so that their relative
influence is governed by the weights rather than by their raw magnitudes,
and mixed with coefficients summing to one. A single scalar exposed to the
outer loop thus controls the trade-off between the two objectives.

\begin{algorithm}[t]
\caption{Gradient combination (expands the update step of
         Algorithm~\ref{alg:anchorgen})}
\label{alg:gradcomb}
\begin{algorithmic}[1]
\Require Current state $\mathbf{z}_{1-\tau}$; decoded pair
         $(\mathbf{z}^{S},\mathbf{z}^{I})$
\Require Gradient projection $\Pi^{S}$; sketch-to-drag ratio $w>0$;
         step size $\eta$; threshold $\epsilon$
\Require Persistent optimizer state $\mathbf{s}$
\State $\lambda\gets 1/(1+w)$
\State $\mathbf{g}^{I}\gets\nabla_{\mathbf{z}_{1-\tau}}
       \mathcal{L}_{\mathrm{anchor}}\!\left(\mathbf{z}^{I}\right)$
\State $\mathbf{g}^{S}\gets\Pi^{S}\,\nabla_{\mathbf{z}_{1-\tau}}
       \mathcal{O}\!\left(\decd^{S}(\mathbf{z}^{S})\right)$
\If{$\|\mathbf{g}^{I}\|_{2}\le\epsilon$ \textbf{and}
    $\|\mathbf{g}^{S}\|_{2}\le\epsilon$}
    \State \textbf{abort}
\EndIf
\State $\mathbf{d}\gets
       (1-\lambda)\,\mathbf{g}^{I}/\|\mathbf{g}^{I}\|_{2}
       +\lambda\,\mathbf{g}^{S}/\|\mathbf{g}^{S}\|_{2}$
\State $(\mathbf{z}_{1-\tau},\mathbf{s})\gets
       \operatorname{AdamStep}\!\left(\mathbf{z}_{1-\tau},\mathbf{d},\mathbf{s};
       \eta\right)$
\State \textbf{return} $\mathbf{z}_{1-\tau}$
\end{algorithmic}
\end{algorithm}

\section{Anchored Optimization on the Manifold}
\label{sec:shapenet_ablation}

This section studies two factors: the form of the target sketch supplied to the
optimizer, and the Sketch/Drag weight ratio $\lambda$ that balances geometric
guidance against the aerodynamic objective. We consider two sketch sources:
(a) an exogenous sketch rendered from a randomly matched reference shape,
and (b) a self sketch rendered from the anchor shape itself, i.e.\ no
external target is provided. Together they measure how much anchor information
survives the optimization.

\parag{Setting.}
We randomly sample $20$ source cases from ShapeNet, retrieve for each source one
target shape whose drag coefficient is $8$--$12\%$ higher and one whose drag
coefficient is $8$--$12\%$ lower, yielding $40$ source--target pairs in total,
and render the side view of every target shape to extract the silhouette that
serves as its target sketch. On this basis we run two experiments.
\textbf{\emph{(a) Exogenous sketch: target reachability.}} The optimizer is
asked to reduce drag by $10\%$ while staying as close as possible to the
prescribed target shape, and we report the shape distance between the optimized
shape and its target.
\textbf{\emph{(b) Self sketch: manifold preservation.}} When no external target
is available, the sketch is taken from the anchor shape itself, which prevents
the optimization from discarding the source identity---for instance, gradually
turning a truck into a sedan. Here we report, over the $20$ source cases, the
shape similarity between each optimized shape and its own source shape,
indicating whether the result remains on the manifold of the original style.

\parag{Results.}
As shown in Fig.~\ref{fig:shapenet_exp}, we report F@$0.5\%$ as a measure of
shape similarity. Each optimization runs for at most $400$ steps and terminates
early once a $10\%$ drag reduction is reached.
In Fig.~\ref{fig:shapenet_exp}(a), where the sketch comes from an exogenous
target shape, a larger sketch weight consistently brings the optimized shape
closer to its target at the point where the $10\%$ drag reduction is attained.
The joint prior supplies a specific deformation direction along which
similarity to the target is maintained while drag decreases, whereas FCSO relies
on a plain shape prior that offers no such directional guidance and therefore
attains a markedly lower similarity. Notably, even when a target sketch
corresponds to a shape that would in principle \emph{increase} drag, the model
still balances the two objectives automatically and finds a viable
drag-descending direction.
In Fig.~\ref{fig:shapenet_exp}(b), where the sketch is rendered from the anchor
itself, increasing the sketch weight initially strengthens the anchoring effect:
more of the source geometry is retained and the result stays on the manifold
better than FCSO. Beyond a certain point, however, the anchor becomes overly
restrictive and confines the optimization to a narrow neighborhood around the
original shape, so drag can no longer descend smoothly and performance degrades.

\begin{figure}[tb]
        \centering
        \includegraphics[width=0.9\linewidth]{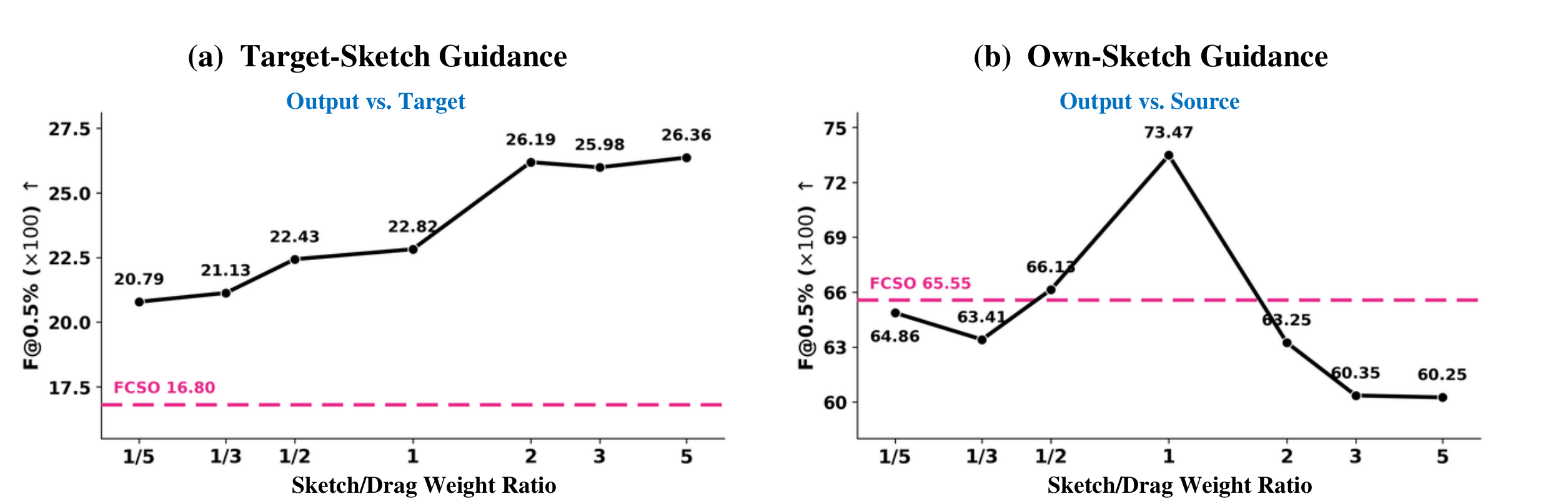}
        \caption{\textbf{Sketch/Drag weight ratio ablation.} F@$0.5\%$
        ($\times 100$, $\uparrow$) versus the Sketch/Drag weight ratio, over the
        $20$ source cases. \textbf{(a)} With an exogenous target sketch, measured
        between the optimized output and the target shape. \textbf{(b)} With the
        anchor's own sketch, measured between the output and its source shape.
        Dashed lines denote the FCSO baseline.}
        \label{fig:shapenet_exp}
\end{figure}

\section{Customized Design Analysis}
\label{sec:customized-design-analysis}

\parag{Diverse Shape Extrapolation}
\label{parag:diverse_shape_extrapolation}

Fig.~\ref{fig:opt_process} shows the shape evolving over 100 optimization epochs under the two objectives, the design intent of adding a rear wing and the drag term. The sketch-to-drag ratio starts at $20{:}1$ and is annealed downward over the course of optimization. Between roughly iterations~40 and~60, the wing begins to form in accordance with the sketch; from~80 to~100, its shape and placement are adjusted to further reduce drag.

\begin{figure}[tb]
	\centering
	\includegraphics[width=0.9\linewidth]{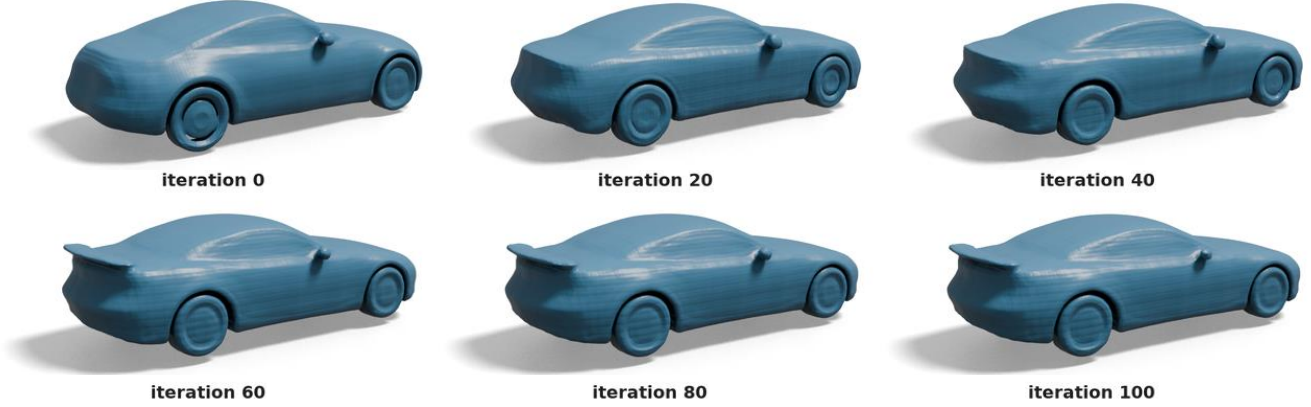}
	\caption{\textbf{Visualization of the Shape over 100 Optimization Iterations for Customized Design.}}
	\label{fig:opt_process}
\end{figure}

\parag{Dim-Err across drag reduction levels.}
\begin{figure}[!htb]
    \centering
    \includegraphics[width=\linewidth]{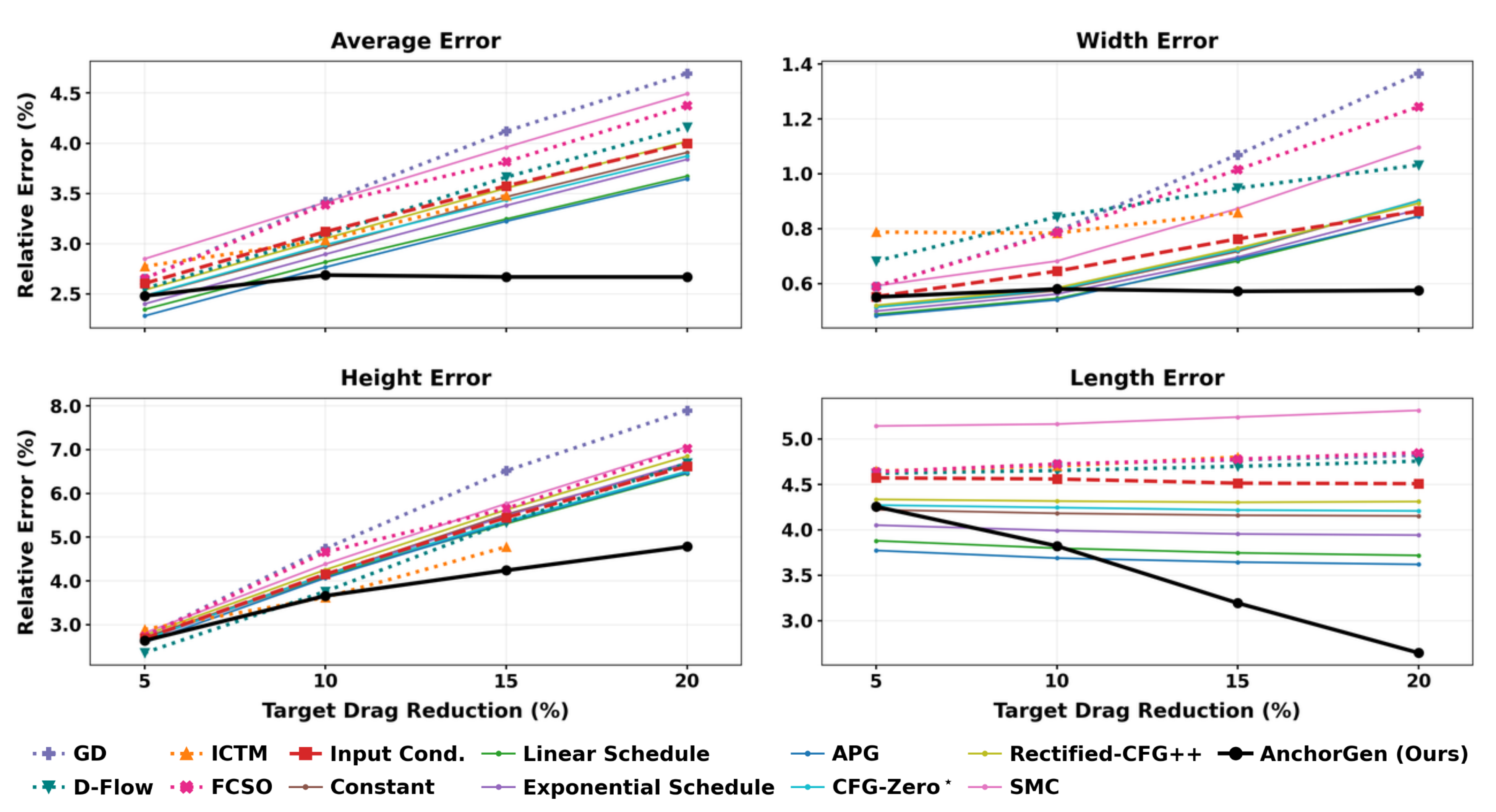}
    \vspace{-0.2in}
    \caption{\textbf{Controlled-dimension error over aerodynamic optimization.}}
    \label{fig:supp_error_curve}
\end{figure}

We report results along the optimization trajectory of
Paragraph~\ref{par:quantified_customized_design}, retaining the $100$-source,
six-target-sketch-per-source DrivAerNet++ setting and examining how target
adherence changes as the aerodynamic objective becomes more demanding. As shown
in Figure~\ref{fig:supp_error_curve}, we evaluate the first valid output whose
realized pressure-drag reduction reaches $5\%$, $10\%$, $15\%$, or $20\%$
relative to its source, where Average Error aggregates the controlled length,
width, and height. All four panels show that our method better preserves the
attributes prescribed by the target sketch, such as the intended length, width,
and height. The side-view length error is the clearest case: as the drag
objective is pushed further, our outputs move \emph{closer} to the target
length rather than away from it. Overall, anchoring the attributes carried by
the target sketch is stronger in AnchorGen than in the CFG-style injection
paradigms, which in turn are stronger than methods that rely on a shape prior
without injecting an explicit directional signal. We also observe that several
baselines become unstable under this finer-grained shape optimization. ICTM,
which independently corrects a partially noised trajectory toward each nominal
drag target, rarely descends all the way to a $20\%$ reduction; FMG, which
conditions a flow on a nominal drag value alone, shows the same difficulty. SGO
enforces no strict manifold projection and is highly prone to shape collapse.

\parag{Shape similarity across drag levels.}

Figure~\ref{fig:supp_shape_curve} shows the complementary full-shape metrics,
averaged over the three views. The trend mirrors that of the controlled
dimensions: as the requested drag reduction grows, every method drifts away
from the shape prescribed by the target sketch, but ours drifts the least. The
condition-only methods degrade steadily on all four metrics, with both F-scores
and normal consistency decreasing while CD-L1 and HD95 increase. CFG variants
partially improve on plain input conditioning, APG forming the strongest
condition-based comparison, yet they follow the same trend. AnchorGen remains
clearly separated from this cluster.

\begin{figure}[!htb]
    \centering
    \includegraphics[width=\linewidth]{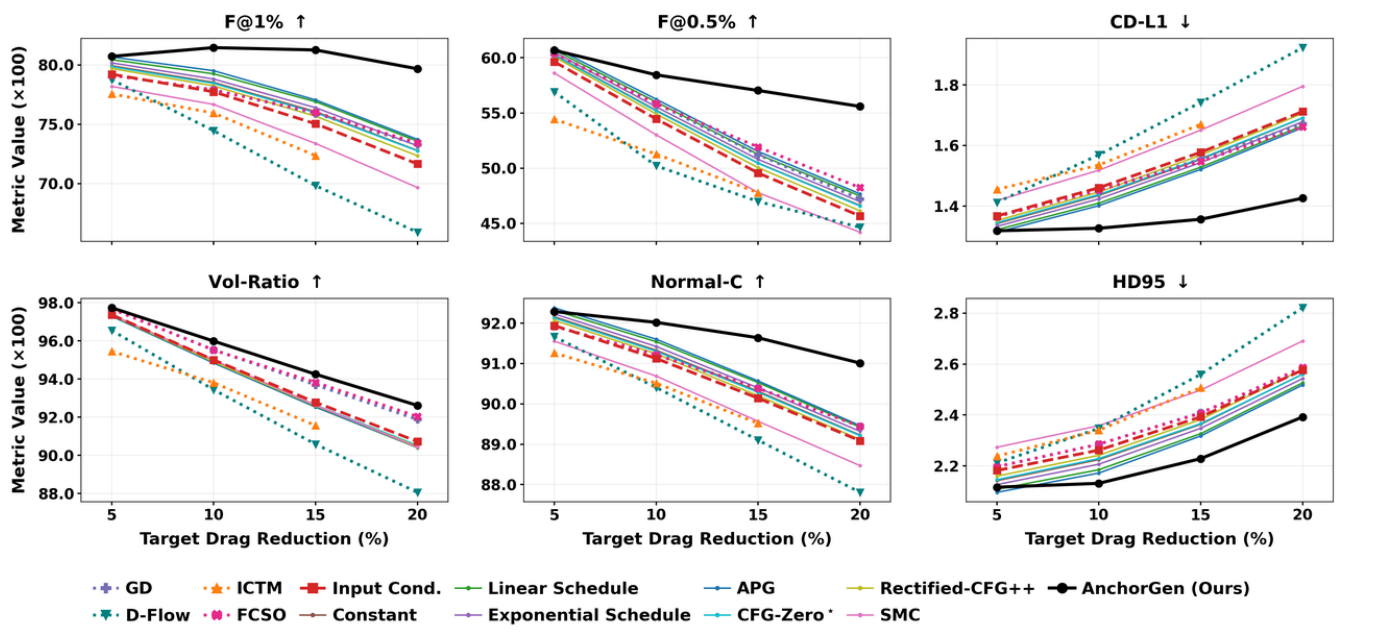}
    \vspace{-0.2in}
    \caption{\textbf{Target-shape similarity over aerodynamic optimization.}}
    \label{fig:supp_shape_curve}
\end{figure}

\parag{Per-view results.}
Tables~\ref{tab:shape_metrics_front}, \ref{tab:shape_metrics_rear}, and
\ref{tab:shape_metrics_side} report the three design dimensions at the $20\%$
operating point. The front-view width task shows the clearest uniform
advantage, and AnchorGen retains a comparable margin on the rear-view height
task, as expected. The side-view length task behaves differently: length is the
dimension least coupled to overall shape, and the matched length variation
across the dataset is correspondingly small, so a change in length does not
translate proportionally into a change in geometry. Some surface metrics
therefore diverge from the controlled-dimension result: FCSO retains higher
global F-scores and a marginally lower CD-L1, whereas AnchorGen attains the
lowest length error and HD95.

  \begin{table}[t]
  \centering
  \caption{\textbf{Evaluation at the $20\%$ target drag reduction for the Front View (Width) setting ($n=200$).} All metrics are scaled by $\times 100$.}
  \label{tab:shape_metrics_front}
  \setlength{\tabcolsep}{4pt}
  \resizebox{\textwidth}{!}{%
  \begin{tabular}{l|ccccccc}
  \toprule
  \textbf{Method}
  & \textbf{F@1\%} $\uparrow$
  & \textbf{F@0.5\%} $\uparrow$
  & \textbf{CD-L1} $\downarrow$
  & \textbf{Vol-Ratio} $\uparrow$
  & \textbf{Normal-C} $\uparrow$
  & \textbf{HD95} $\downarrow$
  & \textbf{W-Err} $\downarrow$ \\
  \midrule
  GD~\citeyearpar{zhan2025physiopt}
  & 64.98 & 44.88 & 1.93 & 92.80 & 87.62 & 2.55 & 1.35 \\

  FMG$^\dagger$~\citeyearpar{Ho22a}
  & NaN & NaN & NaN & NaN & NaN & NaN & NaN \\

  D-Flow~\citeyearpar{BenHamu24}
  & 67.06 & 48.83 & 2.00 & 88.89 & 87.46 & 2.76 & 1.03 \\

  ICTM$^\dagger$~\citeyearpar{Zhang25e}
  & NaN & NaN & NaN & NaN & NaN & NaN & NaN \\

  SGO$^\dagger$~\citeyearpar{chen2026optimization}
  & NaN & NaN & NaN & NaN & NaN & NaN & NaN \\

  FCSO~\citeyearpar{Seiler26a}
  & 64.66 & 44.36 & 1.94 & 92.96 & 87.42 & 2.57 & 1.23 \\
  \midrule
  Input Cond.~\citeyearpar{du2025sgcdiff}
  & 65.99 & 45.56 & 1.91 & 91.61 & 87.66 & 2.54 & 0.86 \\

  Constant~\citeyearpar{xi2024analysis}
  & 68.28 & 47.55 & 1.88 & 90.58 & 87.95 & 2.53 & 0.90 \\

  Linear Schedule~\citeyearpar{xi2024analysis}
  & 69.53 & 49.00 & 1.84 & 90.34 & 88.23 & 2.51 & 0.85 \\

  Exponential Schedule~\citeyearpar{xi2024analysis}
  & 68.99 & 48.19 & 1.86 & 90.48 & 88.11 & 2.52 & 0.87 \\

  APG~\citeyearpar{sadat2024eliminating}
  & 69.57 & 49.08 & 1.84 & 90.36 & 88.26 & 2.51 & 0.84 \\

  CFG-Zero$^\star$~\citeyearpar{fan2025cfg}
  & 68.08 & 47.32 & 1.89 & 90.67 & 87.89 & 2.53 & 0.90 \\

  Rectified-CFG++~\citeyearpar{saini2026rectified}
  & 67.64 & 46.97 & 1.90 & 90.73 & 87.78 & 2.55 & 0.89 \\

  SMC~\citeyearpar{wang2026cfg}
  & 64.12 & 44.24 & 2.03 & 91.17 & 86.76 & 2.66 & 1.10 \\
  \midrule
  AnchorGen (Ours)
  & \textbf{80.26}
  & \textbf{57.02}
  & \textbf{1.41}
  & \textbf{94.75}
  & \textbf{91.31}
  & \textbf{2.35}
  & \textbf{0.58} \\
  \bottomrule
  \end{tabular}%
  }
  \end{table}

\begin{table}[t]
  \centering
  \caption{\textbf{Evaluation at the $20\%$ target drag reduction for the Rear View (Height) setting ($n=200$).} All metrics are scaled by $\times 100$.}
  \label{tab:shape_metrics_rear}
  \setlength{\tabcolsep}{4pt}
  \resizebox{\textwidth}{!}{%
  \begin{tabular}{l|ccccccc}
  \toprule
  \textbf{Method}
  & \textbf{F@1\%} $\uparrow$
  & \textbf{F@0.5\%} $\uparrow$
  & \textbf{CD-L1} $\downarrow$
  & \textbf{Vol-Ratio} $\uparrow$
  & \textbf{Normal-C} $\uparrow$
  & \textbf{HD95} $\downarrow$
  & \textbf{H-Err} $\downarrow$ \\
  \midrule
  GD~\citeyearpar{zhan2025physiopt}
  & 77.50 & 47.17 & 1.48 & 91.43 & 91.57 & \textbf{2.23} & 7.88 \\

  FMG$^\dagger$~\citeyearpar{Ho22a}
  & NaN & NaN & NaN & NaN & NaN & NaN & NaN \\

  D-Flow~\citeyearpar{BenHamu24}
  & 64.83 & 41.82 & 1.83 & 87.65 & 88.80 & 2.51 & 6.74 \\

  ICTM$^\dagger$~\citeyearpar{Zhang25e}
  & NaN & NaN & NaN & NaN & NaN & NaN & NaN \\

  SGO$^\dagger$~\citeyearpar{chen2026optimization}
  & NaN & NaN & NaN & NaN & NaN & NaN & NaN \\

  FCSO~\citeyearpar{Seiler26a}
  & 77.68 & 48.82 & 1.46 & 91.59 & 91.53 & 2.24 & 7.03 \\
  \midrule
  Input Cond.~\citeyearpar{du2025sgcdiff}
  & 75.10 & 45.76 & 1.53 & 90.50 & 90.79 & 2.25 & 6.62 \\

  Constant~\citeyearpar{xi2024analysis}
  & 75.95 & 46.12 & 1.52 & 90.75 & 90.84 & 2.25 & 6.67 \\

  Linear Schedule~\citeyearpar{xi2024analysis}
  & 76.87 & 47.03 & 1.50 & 90.88 & 91.03 & 2.24 & 6.45 \\

  Exponential Schedule~\citeyearpar{xi2024analysis}
  & 76.24 & 46.43 & 1.51 & 90.78 & 90.89 & 2.25 & 6.71 \\

  APG~\citeyearpar{sadat2024eliminating}
  & 77.08 & 47.31 & 1.49 & 90.93 & 91.09 & \textbf{2.23} & 6.47 \\

  CFG-Zero$^\star$~\citeyearpar{fan2025cfg}
  & 75.97 & 46.02 & 1.52 & 90.79 & 90.83 & 2.24 & 6.51 \\

  Rectified-CFG++~\citeyearpar{saini2026rectified}
  & 75.46 & 45.50 & 1.53 & 90.65 & 90.74 & 2.26 & 6.85 \\

  SMC~\citeyearpar{wang2026cfg}
  & 71.19 & 42.74 & 1.63 & 89.95 & 90.03 & 2.29 & 7.06 \\
  \midrule
  AnchorGen (Ours)
  & \textbf{84.26}
  & \textbf{63.01}
  & \textbf{1.27}
  & \textbf{92.63}
  & \textbf{92.55}
  & 2.31
  & \textbf{4.78} \\
  \bottomrule
  \end{tabular}%
  }
  \end{table}

  \begin{table}[t]
  \centering
  \caption{\textbf{Evaluation at the $20\%$ target drag reduction for the Side View (Length) setting ($n=200$).} All metrics are scaled by $\times 100$.}
  \label{tab:shape_metrics_side}
  \setlength{\tabcolsep}{4pt}
  \resizebox{\textwidth}{!}{%
  \begin{tabular}{l|ccccccc}
  \toprule
  \textbf{Method}
  & \textbf{F@1\%} $\uparrow$
  & \textbf{F@0.5\%} $\uparrow$
  & \textbf{CD-L1} $\downarrow$
  & \textbf{Vol-Ratio} $\uparrow$
  & \textbf{Normal-C} $\uparrow$
  & \textbf{HD95} $\downarrow$
  & \textbf{L-Err} $\downarrow$ \\
  \midrule
  GD~\citeyearpar{zhan2025physiopt}
  & 77.42 & 49.60 & 1.61 & 91.42 & 89.14 & 2.95 & 4.83 \\

  FMG$^\dagger$~\citeyearpar{Ho22a}
  & NaN & NaN & NaN & NaN & NaN & NaN & NaN \\

  D-Flow~\citeyearpar{BenHamu24}
  & 65.83 & 43.29 & 1.94 & 87.65 & 87.16 & 3.20 & 4.76 \\

  ICTM$^\dagger$~\citeyearpar{Zhang25e}
  & NaN & NaN & NaN & NaN & NaN & NaN & NaN \\

  SGO$^\dagger$~\citeyearpar{chen2026optimization}
  & NaN & NaN & NaN & NaN & NaN & NaN & NaN \\

  FCSO~\citeyearpar{Seiler26a}
  & \textbf{77.97}
  & \textbf{51.54}
  & \textbf{1.58}
  & \textbf{91.58}
  & \textbf{89.38}
  & 2.96
  & 4.85 \\
  \midrule
  Input Cond.
  & 73.92 & 45.71 & 1.69 & 90.07 & 88.82 & 2.94 & 4.51 \\

  Constant~\citeyearpar{xi2024analysis}
  & 74.16 & 46.21 & 1.67 & 90.06 & 88.89 & 2.90 & 4.15 \\

  Linear Schedule~\citeyearpar{xi2024analysis}
  & 74.36 & 46.43 & 1.66 & 89.98 & 89.00 & 2.83 & 3.72 \\

  Exponential Schedule~\citeyearpar{xi2024analysis}
  & 74.36 & 46.31 & 1.66 & 90.03 & 88.96 & 2.87 & 3.94 \\

  APG~\citeyearpar{sadat2024eliminating}
  & 74.55 & 46.65 & 1.65 & 90.04 & 89.04 & 2.81 & 3.62 \\

  CFG-Zero$^\star$~\citeyearpar{fan2025cfg}
  & 74.26 & 46.38 & 1.67 & 90.10 & 88.92 & 2.91 & 4.21 \\

  Rectified-CFG++~\citeyearpar{saini2026rectified}
  & 73.97 & 45.95 & 1.68 & 90.03 & 88.83 & 2.93 & 4.31 \\

  SMC~\citeyearpar{wang2026cfg}
  & 73.71 & 45.63 & 1.73 & 90.16 & 88.61 & 3.13 & 5.32 \\
  \midrule
  AnchorGen (Ours)
  & 74.54
  & 46.76
  & 1.60
  & 90.47
  & 89.18
  & \textbf{2.51}
  & \textbf{2.65} \\
  \bottomrule
  \end{tabular}%
  }
  \end{table}

\parag{Visualization.}
Figs.~\ref{fig:supp_front}, \ref{fig:supp_rear}, and \ref{fig:supp_side} show,
from the front, rear, and side views, how well the different methods hold the
anchor during optimization. Our paradigm anchors width, height, and length more
tightly once the full $20\%$ drag reduction is reached.
\begin{figure}[!htb]
    \centering
    \includegraphics[width=\linewidth]{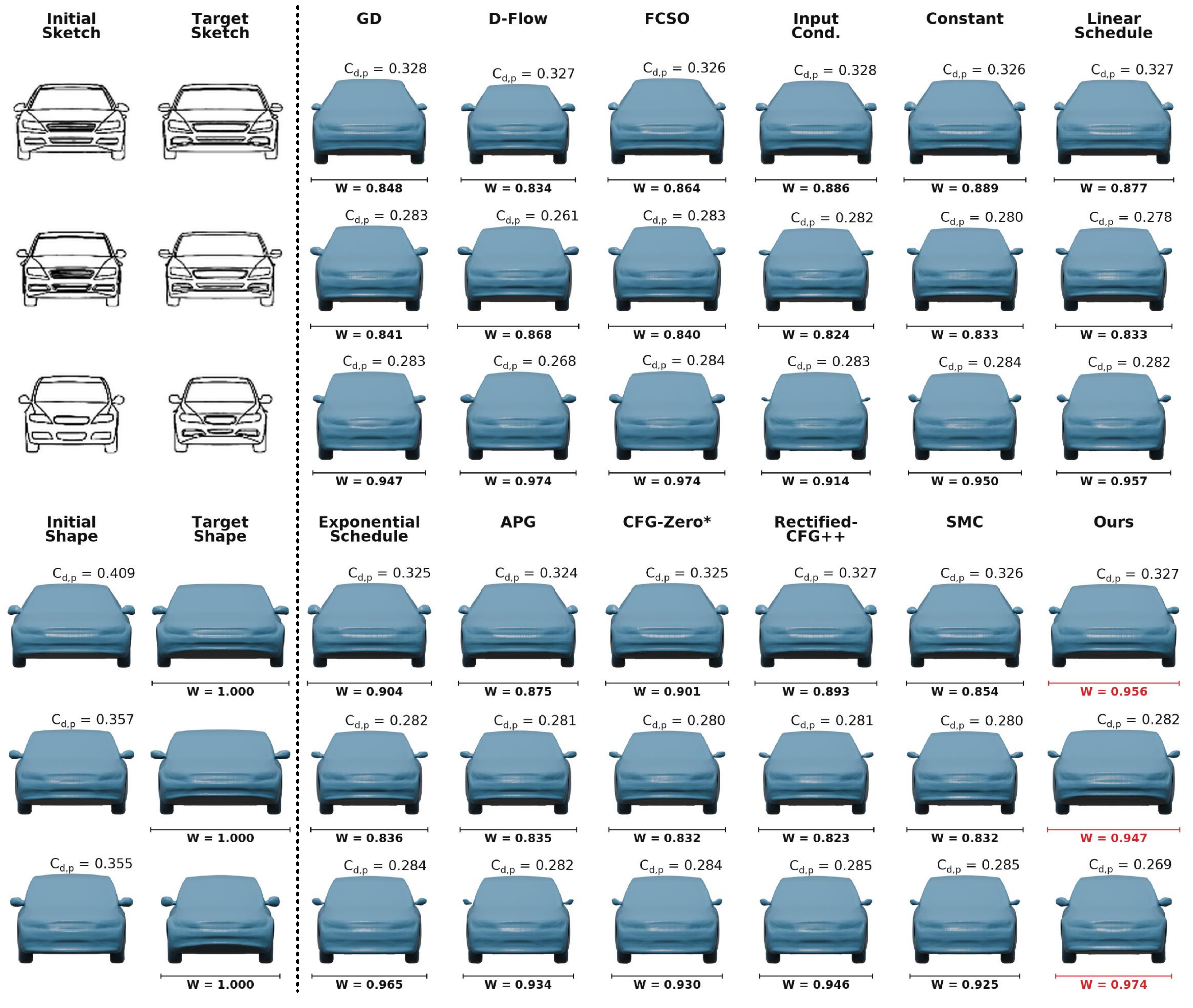}
    \vspace{-0.2in}
    \caption{\textbf{Front view (width).} Effect of the target sketch anchor.
    The width of the shape corresponding to the target sketch is taken as $1$,
    and we report the ratio to that width reached after a $20\%$ drag
    reduction.}
    \label{fig:supp_front}
\end{figure}

\begin{figure}[!htb]
    \centering
    \includegraphics[width=\linewidth]{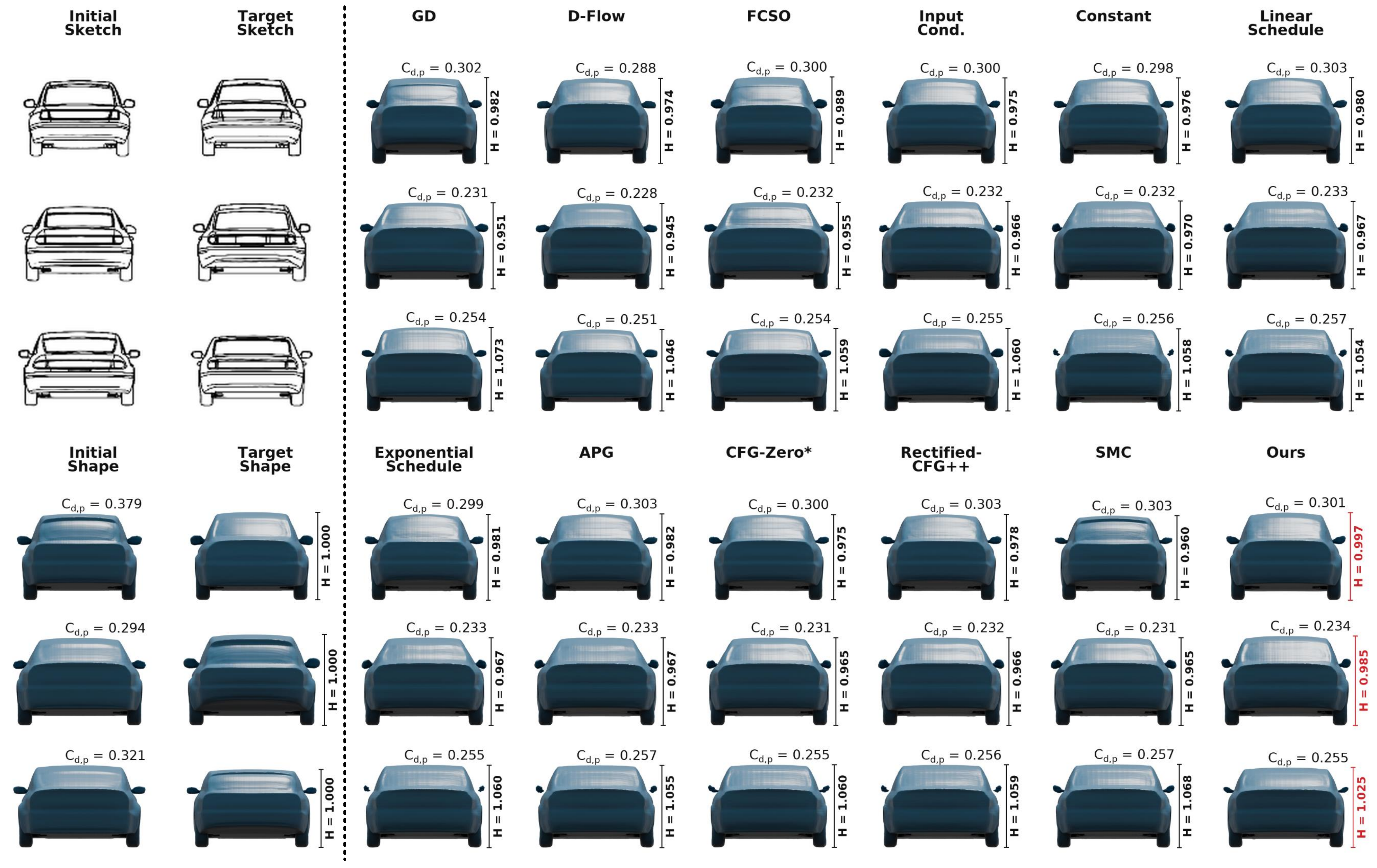}
    \vspace{-0.2in}
    \caption{\textbf{Rear view (height).} Effect of the target sketch anchor.
    The height of the shape corresponding to the target sketch is taken as $1$,
    and we report the ratio to that height reached after a $20\%$ drag
    reduction.}
    \label{fig:supp_rear}
\end{figure}

\clearpage
\begin{figure}[p]
    \centering
    \includegraphics[
        width=\linewidth,
        height=\dimexpr\textheight-4\baselineskip-\abovecaptionskip-\belowcaptionskip\relax,
        keepaspectratio
    ]{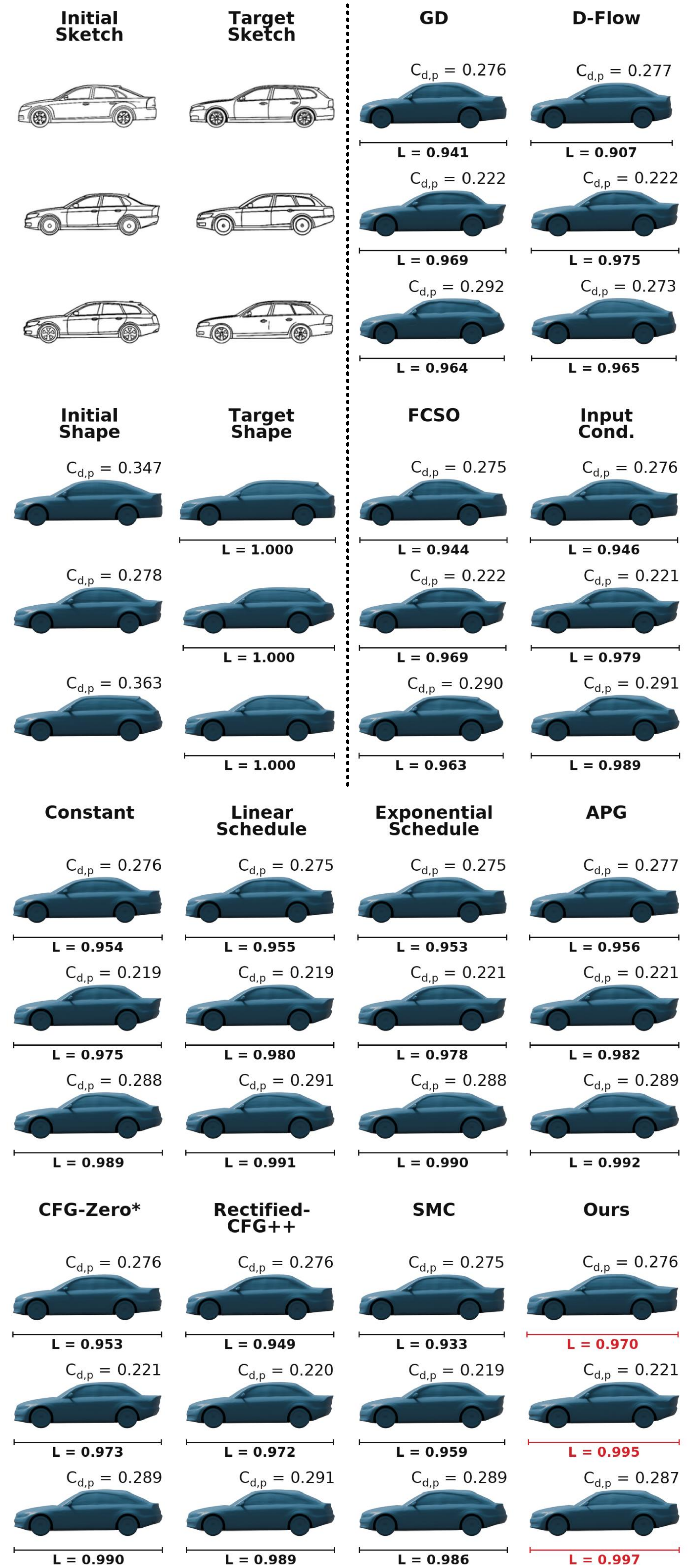}
    \caption{\textbf{Side view (length).} Effect of the target sketch anchor.
    The length of the shape corresponding to the target sketch is taken as $1$,
    and we report the ratio to that length reached after a $20\%$ drag
    reduction.}
    \label{fig:supp_side}
\end{figure}
\clearpage

\section{Ablation}
\label{sec:ablation}

\parag{Ablation on the inversion depth $\tau$.}
We ablate the inversion depth $\tau$ (Algorithm~\ref{alg:anchorgen}) on the $200$ cases of the Front View (Width) setting to study how far the initial latent should be inverted before optimization. As shown in Table~\ref{tab:ablation_tau_front}, a shallow inversion ($\tau=0.1$ and $0.2$) makes it difficult to anchor the optimized shape to the target sketch, yielding an F@1\% of only $66.51$ and $72.14$. Increasing the depth to $\tau=0.4$ and $0.5$ brings substantial improvements, raising F@1\% to $80.26$ and $82.30$ and reducing CD-L1 to $1.41$ and $1.35$, respectively. However, a deeper inversion requires integrating and back-propagating through a longer segment of the flow at every iteration. To balance performance and efficiency, we set $\tau=0.4$ on DrivAerNet++.
\begin{table}[t]
\centering
\caption{\textbf{Ablation on the inversion depth $\tau$ (Algorithm~\ref{alg:anchorgen}) at the $20\%$ target drag reduction for the Front View (Width) setting.} All metrics are scaled by $\times 100$.}
\label{tab:ablation_tau_front}
\setlength{\tabcolsep}{4pt}
\resizebox{\textwidth}{!}{%
\begin{tabular}{l|ccccccc}
\toprule
$\boldsymbol{\tau}$
& \textbf{F@1\%} $\uparrow$
& \textbf{F@0.5\%} $\uparrow$
& \textbf{CD-L1} $\downarrow$
& \textbf{Vol-Ratio} $\uparrow$
& \textbf{Normal-C} $\uparrow$
& \textbf{HD95} $\downarrow$
& \textbf{W-Err} $\downarrow$ \\
\midrule
$0.1$
& 66.51 & 45.22 & 1.85 & 93.59 & 88.23 & 2.45 & 1.23 \\

$0.2$
& 72.14 & 48.62 & 1.64 & 94.53 & 89.62 & 2.33 & 0.81 \\

$0.3$
& 76.78 & 52.56 & 1.50 & 94.70 & 90.69 & \textbf{2.30} & 0.62 \\

$0.4$
& 80.26 & 57.02 & 1.41 & \textbf{94.75} & 91.31 & 2.35 & \textbf{0.58} \\

$0.5$
& \textbf{82.30} & \textbf{61.95} & \textbf{1.35} & 94.64 & \textbf{91.67} & 2.49 & 0.62 \\
\bottomrule
\end{tabular}%
}
\end{table}

\parag{Model-capacity scaling.}

We vary Transformer depth to scale the parameter count of the rectified-flow
models. As summarized in Table~\ref{tab:anchorgen_capacity}, depths
$\{2,5,7,10,12\}$ correspond to approximately
$\{14.06,34.54,48.19,68.67,82.32\}$M parameters. At each depth we compare
AnchorGen against a shape-only FCSO refiner of nearly identical capacity: the
two differ by only $1{,}024$ parameters and are therefore indistinguishable
when reported to two decimals in millions. The depth-2, depth-5, and depth-7
AnchorGen models share the same $10{,}000$-step protocol, batch size $128$, and
learning rate $10^{-4}$. The depth-10 model uses batch size $64$ and learning
rate $5\times10^{-5}$, as the matched configuration exceeded available memory,
and the depth-12 points reuse the existing full-size references. The five-point
curves should thus be read as empirical capacity trends, with depths 2, 5, and
7 forming the strictly matched-protocol core of the ablation.

\begin{table}[t]
  \centering
  \caption{AnchorGen model capacity variants.}
  \label{tab:anchorgen_capacity}
  \setlength{\tabcolsep}{10pt}
  \begin{tabular}{lcc}
  \toprule
  \textbf{Variant}
  & \textbf{Transformer Depth}
  & \textbf{Parameters (M)} \\
  \midrule
  P14-D2  & 2  & 14.06 \\
  P35-D5  & 5  & 34.54 \\
  P48-D7  & 7  & 48.19 \\
  P69-D10 & 10 & 68.67 \\
  P82-D12 & 12 & 82.32 \\
  \bottomrule
  \end{tabular}
  \end{table}

We retain the quantified customized-design protocol of
Paragraph~\ref{par:quantified_customized_design}: each of $100$ DrivAerNet++
sources is paired with two targets along length, width, and height, giving
$600$ frozen source--target pairs ($200$ per view). Every variant uses the same
pairs, a budget of at most $800$ iterations, and learning rate $0.03$; we
evaluate the first valid checkpoint reaching at least $20\%$ pressure-drag
reduction. AnchorGen runs one target-sketch-conditioned optimization per pair,
whereas FCSO produces one target-agnostic shape per source, evaluated against
its six assigned targets. Figure~\ref{fig:supp_param} reports the same seven
target-shape metrics and $\times100$ scaling as
Table~\ref{tab:total_metrics_scaled}.

Increasing AnchorGen capacity improves the overall target-adherence envelope,
with most of the gain already realized at mid-sized models. From $14.06$M to
$82.32$M parameters, F@$0.5\%$ rises from $51.26$ to $55.60$ and average
controlled-dimension error falls from $3.61$ to $2.67$; the individual surface
metrics are not strictly monotonic and largely saturate beyond $34.54$M. The
capacity-matched FCSO variants, by contrast, stay nearly flat on all seven
metrics. More importantly, the smallest AnchorGen model already outperforms the
largest FCSO model in every reported column, for instance $51.26$ against
$48.24$ in F@$0.5\%$ and $3.61$ against $4.37$ in dimension error. Capacity
alone therefore does not recover the missing design direction: the explicit
target-sketch anchor is effective at the smallest tested capacity, and added
capacity mainly strengthens its ability to retain that direction under
aerodynamic optimization.
\begin{figure}[!htb]
    \centering
    \includegraphics[width=\linewidth]{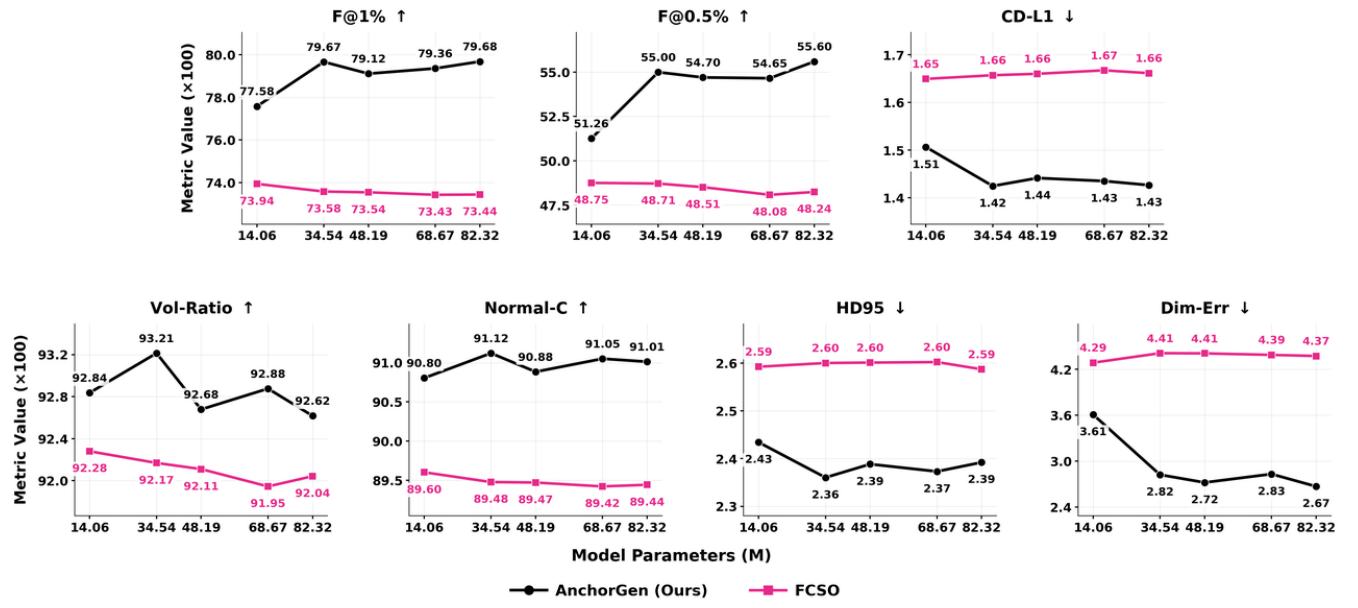}
    \vspace{-0.2in}
    \caption{\textbf{Target-shape metrics at $20\%$ drag reduction across AnchorGen and FCSO model capacities.}}
    \label{fig:supp_param}
\end{figure}

\parag{Diversity of controlled edits.}
\label{par:diversity_of_controlled_edits}

Correcting with a shape prior alone, as in FCSO, tends to pull the optimization
toward the high-density region of the learned distribution, leaving little room
for explicit directional control. Supplying the direction explicitly during
optimization avoids this collapse. We visualize $50$ cases spanning different
views and edit directions, including lengthening and shortening in the side
view and widening and narrowing in the front view; relative to their sources,
these edits should ideally be diverse.

For each source we encode its decoded source mesh together with the AnchorGen
and FCSO outputs at the $20\%$ operating point using the same frozen VecSet
encoder. Each mesh is represented by four deterministic $2{,}048$-point surface
samples, and the encoder posterior mean gives $256$ unordered eight-channel
tokens per sample. To avoid imposing a spurious token correspondence, we pool
the four token sets and compute seven permutation-invariant marginal statistics
per channel---mean, standard deviation, and the $10$th, $25$th, $50$th, $75$th,
and $90$th percentiles---yielding a $56$-D feature $p(x)$. For method
$m\in\{\mathrm{AnchorGen},\mathrm{FCSO}\}$ and source $i$, the visualized edit
is the source-centered difference $\Delta_i^m=p(x_i^m)-p(s_i)$ rather than the
raw optimized-shape feature. As shown in Figure~\ref{fig:supp_pca}, our edits spread more widely, whereas the
shape-prior baseline concentrates on a few high-density modes.
\begin{figure}[!htb]
    \centering
    \includegraphics[width=\linewidth]{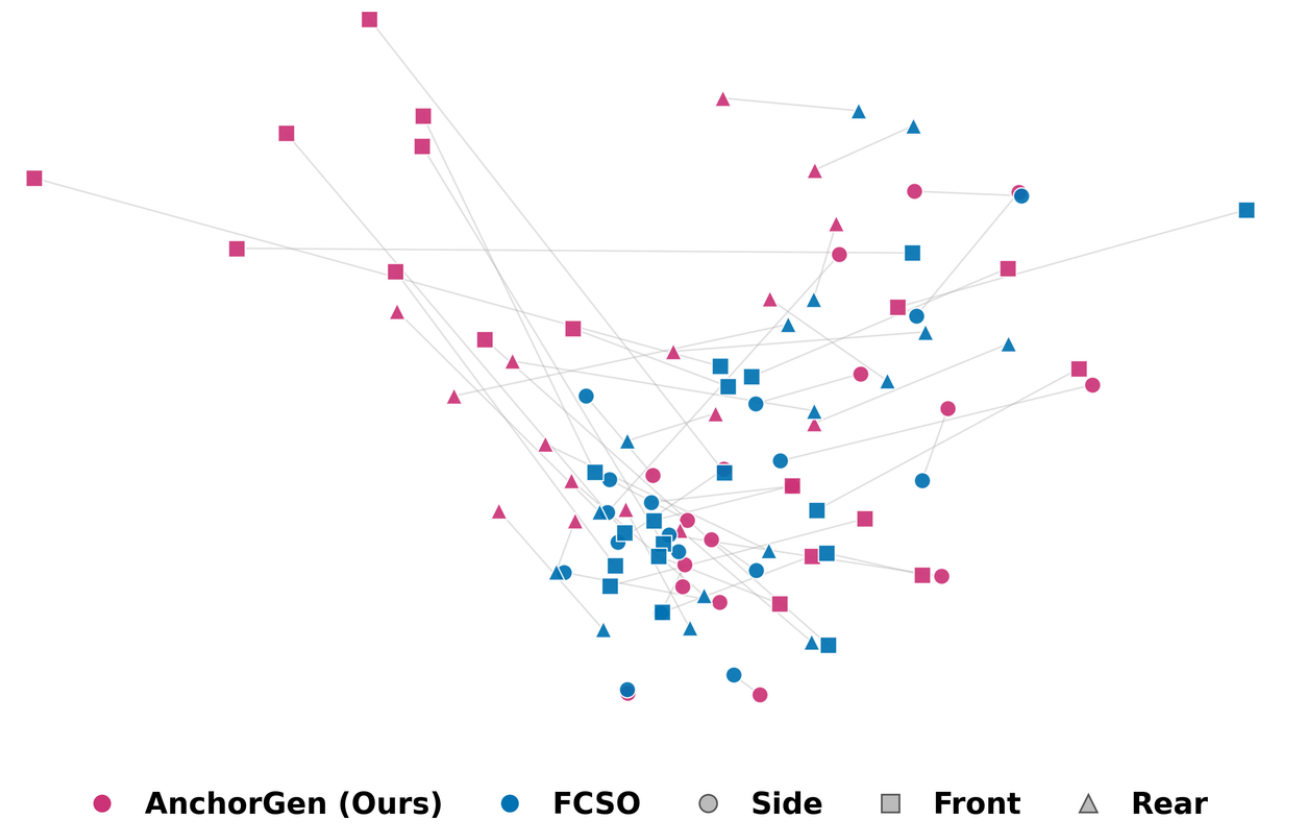}
    \vspace{-0.2in}
    \caption{\textbf{Shared PCA of source-centered VecSet edit features for matched AnchorGen and FCSO outputs.}}
    \label{fig:supp_pca}
\end{figure}

\section{Generation from Scratch}
\label{sec:generation_from_scratch}

\paragraph{Setting.}
We compare against PhysGen~\citep{You26a} using its official code and released weights.
Each PhysGen cycle has two stages: the clean Dora latent is first refined with 30 Adam steps
(learning rate $5\times10^{-3}$, proximity regularizer of weight $10^{-3}$ toward the cycle's starting latent),
then re-noised to $\sigma\approx0.25$ and denoised with 75 Euler steps of sketch-conditioned sampling (CFG scale 5)., and we run 60 cycles.
AnchorGen also runs 60 iterations, each consisting of a single drag evaluation and a single Adam update
(learning rate $0.03$) of the joint flow state, for a total of 60 surrogate queries. In terms of model size, AnchorGen has $138.20$M parameters ($82.32$M flow-matching model + $55.88$M VAE), whereas PhysGen has $399.59$M ($208.00$M diffusion model + $191.59$M VAE).

\paragraph{CFD simulation.}
Both methods optimize against the surrogate; we evaluate the final meshes with OpenFOAM~9~\cite{jasak2007openfoam}.
We solve the steady incompressible RANS equations with \texttt{simpleFoam} and the $k$--$\omega$ SST turbulence model,
with a freestream of $30\,\mathrm{m/s}$ along $+x$, $\rho=1.205\,\mathrm{kg/m^3}$, and $\nu=1.588\times10^{-5}\,\mathrm{m^2/s}$.
The domain spans $x\in[-12,32]$, $y\in[-4,4]$, $z\in[-0.3,6.4]\,\mathrm{m}$, with a moving ground at $z=-0.3\,\mathrm{m}$,
a stationary car wall, and slip lateral and top boundaries.
Meshes receive only minimal repairs (merging duplicate vertices, removing degenerate faces, fixing normals),
and volume meshes are generated with \texttt{snappyHexMesh} using the same refinement recipe for all methods.
After \texttt{potentialFoam} initialization, we run 2500 iterations without early termination and report the pressure drag coefficient
$C_d = F_x^{\mathrm{p}} / (\tfrac12\rho U^2 A_{\mathrm{ref}})$, where $F_x^{\mathrm{p}}$ is the pressure (not total) component of the streamwise force on the car, averaged over iterations 2000--2500,
with $A_{\mathrm{ref}}$ fixed per case and shared across the target and all methods.
We compute the per-case relative reduction $(C_d^{\mathrm{target}}-C_d^{\mathrm{method}})/C_d^{\mathrm{target}}$
and average it with equal weight over the 10 cases.
A run is flagged as not converged if the two halves of the averaging window differ in mean by more than $2\%$
or if the standard deviation exceeds $5\%$ of the mean.

\end{document}